%% file: AnonymousSubmission2027.tex
\documentclass[letterpaper]{article} 
\usepackage{aaai2027}  
\usepackage[hyphens]{url}  
\usepackage{graphicx} 
\usepackage{natbib}  
\usepackage{caption} 
\usepackage{algorithm}
\usepackage{algorithmic}

\usepackage{newfloat}
\usepackage{listings}
\DeclareCaptionStyle{ruled}{labelfont=normalfont,labelsep=colon,strut=off} 
\floatstyle{ruled}
\newfloat{listing}{tb}{lst}{}
\floatname{listing}{Listing}

\usepackage{booktabs}
\usepackage{dashrulex}

\usepackage{tcolorbox} 

\usepackage{amsmath, amsthm}

\usepackage{xcolor}     
\usepackage{colortbl}   
\usepackage{graphicx}  
\usepackage{subcaption} 
\usepackage{multirow}
\usepackage{multicol}
\usepackage{amssymb}

\usepackage{nicematrix}

\usepackage{tabularx} 
\usepackage{array}
\usepackage{amsfonts}
\usepackage{bm}

\definecolor{lightgraybg}{gray}{0.93}
\definecolor{mysoftgreen}{RGB}{232, 245, 233}  
\definecolor{mysoftblue}{RGB}{235, 243, 250}   
\definecolor{mydarkgreen}{rgb}{0.0, 0.5, 0.0} 
\definecolor{graytext}{gray}{0.6}

\newcommand{\confyear}[1]{\textcolor{graytext}{\texttt{\scriptsize(#1)}}}

\usepackage{multibib}
\newcites{app}{Appendix References}

\title{Centering before Pruning: Lightweight Geometry Correction \\ for Diversity-Based Visual Token Pruning in LVLMs}
\author{
    Shunjie Wen,
    Jaeyeon Lee, 
    Dong-Wan Choi\corresponding
}
\affiliations{
    Inha University

}

\begin{document}

\maketitle

\begin{abstract}

Large vision-language models (LVLMs) incur substantial inference costs due to their long and highly redundant visual-token sequences. Diversity-based pruning mitigates this cost by selecting token subsets based on pairwise cosine similarity. We find, however, that similarities between raw visual tokens are strongly concentrated in the positive range, limiting their ability to distinguish non-redundant tokens. A natural way to improve this resolution is to center token features before computing cosine similarity. Centering indeed reveals a substantially richer pairwise structure, yet unexpectedly degrades pruning performance when used alone. We show that this apparent contradiction arises because the raw geometry does more than represent pairwise diversity: it also implicitly favors globally distinctive tokens, which tend to contain semantically informative content. Centering better resolves subset diversity but loses this useful token-wise preference, revealing that diversity and distinctiveness are entangled in the raw geometry. Based on this analysis, we propose the \textbf{Cen}tered Geometry \textbf{Prune}r (Cen-Prune), which measures subset diversity using centered cosine similarity while retaining raw-space distinctiveness as a complementary token-wise preference. This lightweight, plug-and-play correction leaves the underlying selection mechanism unchanged and incurs negligible computational overhead. Extensive experiments across multiple image- and video-understanding benchmarks and LVLM architectures demonstrate that Cen-Prune provides robust improvements in overall performance across existing diversity-based pruners.

\end{abstract}


\input{Section_1}
\input{Section_2}
\input{Section_3}
\input{Section_4}

\input{Section_5}
\input{Section_6}
\bibliography{aaai2027}

\newpage
\input{Appendix}




\bibliographystyleapp{aaai2027} 
\bibliographyapp{aaai2027}     


\end{document}

%% file: Section_1.tex
\section{Introduction}
Large vision-language models (LVLMs)~\cite{llava/liu2023visual,llavanext/liu2024llavanext,llava-video/ZhangWLLMLL25, qwen2.5/bai2025qwen25vltechnicalreport} achieve strong multimodal understanding but incur substantial inference costs. Images are represented as dense grids of patch tokens~\cite{Vit/DosovitskiyB0WZ21}, and recent models further expand these sequences through dynamic-resolution processing for high-resolution images~\cite{llavanext/liu2024llavanext,InternVL3/zhu2025internvl3, qwen2.5/bai2025qwen25vltechnicalreport} and multi-frame processing for videos~\cite{LLaVA-OneVision/0080ZGZ00ZZL0L25,llava-video/ZhangWLLMLL25}. Consequently, visual tokens can occupy a substantial portion of the sequence processed by the language model. As attention cost grows quadratically with sequence length, visual tokens become a major source of latency and memory consumption~\cite{SparseVLM/zhang2025sparsevlm}, motivating the reduction of visual tokens for efficient LVLM inference.



Visual token pruning offers a practical way to reduce this cost~\cite{SparseVLM/zhang2025sparsevlm}. Since visual representations contain substantial redundancy, many tokens can be removed with limited performance degradation. Training-free methods identify the tokens to retain using different criteria. Attention-based methods rank tokens using attention from the vision encoder or the language model~\cite{VisionZip/YangCTWL0J25,HoloV/abs-2510-02912,VisPruner/ZhangCLZZCGSZ25}. Diversity-based methods instead retain mutually dissimilar tokens, either through purely geometric~\cite{DivPrune/AlvarSAZ25}, instruction-conditioned~\cite{CDPruner/zhang2026beyond}, or sensitivity-guided selection~\cite{ZOO-Prune/Kim_2026_CVPR}. Other methods suppress redundancy through semantic grouping~\cite{PruneSID/fang2026prune} or multimodal coverage~\cite{MMTok/dong2026mmtok}. Despite these variations, diversity-based selection fundamentally relies on pairwise cosine similarity to measure visual-token differences.


Yet, the geometry underlying this similarity has remained largely unexplored~\cite{WoldLike/lee2026more}. Diversity-based pruning assumes that the vision encoder provides a proper embedding space to distinguish redundant from non-redundant tokens. We find, however, that pairwise similarities among visual tokens are strongly concentrated in the positive range, making most tokens appear mutually similar. Existing selectors can still suppress near-duplicates, but the resulting subsets remain composed of positively similar tokens. Interestingly, this restricted geometry is not entirely uninformative: it implicitly favors globally distinctive tokens, which are often semantically important. Thus, existing diversity-based pruning operates on a geometry that retains an implicit importance signal but provides limited resolution for subset diversity.


Feature centering appears to provide an intuitive remedy. It exposes the missing diversity by removing the shared component of visual tokens, and spreads pairwise similarities into a substantially broader range and reveals centroid-relative directions that are obscured in the raw geometry. Nevertheless, directly applying existing selectors to this centered geometry reduces downstream performance. Centering removes the implicit importance preference of the raw geometry, allowing tokens to be selected primarily because their residual directions differ from the current subset, regardless of whether those variations are informative. This reveals a fundamental tension: the raw geometry preserves token importance but poorly resolves diversity, whereas the centered geometry reveals diversity but cannot determine which diverse tokens are worth retaining.


Motivated by this observation, we propose \textbf{Cen-Prune}, a lightweight, plug-and-play framework that explicitly unites importance and diversity for visual token pruning. Cen-Prune assigns the two roles to complementary geometries: it constructs the pairwise similarity matrix from centered features to capture subset diversity, while assessing token importance through angular distinctiveness in the raw feature geometry. Specifically, we introduce a raw-space distinctiveness score derived from raw cosine similarities and incorporate it into the original selection rule without changing the underlying pruning procedure. Across diverse image- and video-understanding benchmarks, Cen-Prune provides robust performance gains for existing diversity-based pruners on multiple LVLM architectures with negligible computational overhead.


%% file: Section_2.tex
\section{Related Works}

Visual token pruning (VTP) reduces the inference cost of large vision-language models by retaining a compact set of visual tokens. Existing methods can be broadly understood through two complementary selection principles: \emph{token importance} and \emph{subset diversity}.


\paragraph{Importance-Based Visual Token Pruning.}
Importance-based methods estimate the contribution of each visual token using signals such as attention, instruction relevance, or semantic coverage.
VisPruner~\cite{VisPruner/ZhangCLZZCGSZ25} estimates token importance from visual attention and subsequently removes redundancy based on feature similarity.
VisionZip~\cite{VisionZip/YangCTWL0J25} retains dominant tokens while merging less important tokens into contextual representations according to semantic similarity.
HoloV~\cite{HoloV/abs-2510-02912} preserves holistic visual evidence through crop-wise adaptive token allocation, while MMTok~\cite{MMTok/dong2026mmtok} formulates selection as a multimodal maximum-coverage problem over textual and visual tokens.
These approaches primarily determine which visual evidence should be preserved, although several also incorporate redundancy reduction to avoid repeatedly retaining similar content.

\paragraph{Diversity-Based Visual Token Pruning.}
Diversity-based methods consider relationships among tokens to retain a complementary subset rather than ranking each token independently.
PruneSID~\cite{PruneSID/fang2026prune} clusters semantically related tokens using Principal Semantic Components Analysis (PSCA) and suppresses redundancy within each group through Non-Maximum Suppression (NMS).
DivPrune~\cite{DivPrune/AlvarSAZ25} more explicitly formulates token selection as a Max--Min Diversity Problem (MMDP), while CDPruner~\cite{CDPruner/zhang2026beyond} combines determinantal diversity with image--text relevance for query-aware pruning.
ZOO-Prune~\cite{ZOO-Prune/Kim_2026_CVPR} further couples a diversity objective with zeroth-order sensitivity estimation to retain influential yet complementary tokens.

\paragraph{Diversity--Importance Entanglement.}
Diversity-based methods, however, are not independent of token importance.
CDPruner~\cite{CDPruner/zhang2026beyond} and ZOO-Prune~\cite{ZOO-Prune/Kim_2026_CVPR} explicitly incorporate relevance or sensitivity scores.
More importantly, our empirical analysis shows that even DivPrune~\cite{DivPrune/AlvarSAZ25}, despite having no explicit importance term, favors tokens that are distinctive relative to the remaining tokens in the image. Such tokens not only increases subset diversity but are also often visually distinctive and informative.
Thus, part of the benefit attributed to diversity may arise from a token-wise preference already encoded in the raw similarity geometry. Prior work has not clearly separated this preference from subset-level diversity, which is the focus of our analysis.

%% file: Section_3.tex
\section{Preliminaries}


This section formulates visual token pruning and two representative diversity-based selection objectives that form the basis of our method.

\paragraph{Visual Token Pruning Problem.}
Given an image, the visual encoder of an LVLM produces a sequence of visual tokens $\bm{V}=[\boldsymbol{v}_1,\ldots,\boldsymbol{v}_N]^\top\in\mathbb{R}^{N\times D}$, which is fed into the language model together with the text tokens.
Visual token pruning aims to select an index set $T\subseteq[N]$ with $|T|=n\ll N$ such that retaining only $\bm{V}_T$ minimizes the degradation in downstream task performance, while reducing the visual sequence length to be processed by the language model.
For a token $\boldsymbol{v}_i$, we define its $\ell_2$-normalized representation as $\hat{\boldsymbol{v}}_i=\boldsymbol{v}_i/\lVert\boldsymbol{v}_i\rVert_2$ and its centered representation as $\bar{\boldsymbol{v}}_i=\boldsymbol{v}_i-\boldsymbol{\mu}$, where $\boldsymbol{\mu}=\frac{1}{N}\sum_{j=1}^{N}\boldsymbol{v}_j$ is the mean token of the corresponding image.
We use the same accents for the associated token matrices: $\hat{\boldsymbol{V}}=[\hat{\boldsymbol{v}}_1,\ldots,\hat{\boldsymbol{v}}_N]^\top$ and $\bar{\boldsymbol{V}}=[\bar{\boldsymbol{v}}_1,\ldots,\bar{\boldsymbol{v}}_N]^\top$.



\paragraph{Diversity-Based Selection.}
To minimize the performance drop after token pruning, diversity-based methods select a subset of mutually dissimilar visual tokens, treating pairwise dissimilarity as a surrogate for preserving non-redundant visual information. More formally, let
$\bm{S}=\hat{\bm{V}}\hat{\bm{V}}^\top$ denote the cosine-similarity matrix, where
$S_{ij}=\langle\hat{\boldsymbol{v}}_i,\hat{\boldsymbol{v}}_j\rangle$, and let $D_{ij}=1-S_{ij}$ be the corresponding distance.
The pruning problem is then formulated as selecting a subset $T\subseteq[N]$ with $|T|=n$ that maximizes diversity under this pairwise geometry. Existing diversity-based visual token pruning methods mainly follow two representative selection mechanisms: \textit{max--min diversity} and \textit{determinantal diversity}.

\paragraph{Max--Min Diversity.}
The max--min diversity objective, adopted by DivPrune~\cite{DivPrune/AlvarSAZ25} and ZOO-Prune~\cite{ZOO-Prune/Kim_2026_CVPR}, maximizes the minimum pairwise distance among the selected tokens:
\begin{equation}
T^\star=\arg\max_{|T|=n}\ \min_{\substack{i,j\in T\\ i\neq j}} D_{ij}.
\end{equation}
This objective is typically approximated by greedy farthest-point selection, which iteratively adds the token that maximizes its minimum distance to the already selected set.

\paragraph{Determinantal Diversity.}
The determinantal diversity objective, adopted by CDPruner~\cite{CDPruner/zhang2026beyond}, favors subsets that span a large volume:
\begin{equation}
T^\star=\arg\max_{|T|=n}\ \det(\bm{L}_T),
\quad
\bm{L}=\operatorname{diag}(\bm{q})\,\bm{S}\,\operatorname{diag}(\bm{q}),
\end{equation}
where $q_i$ denotes the quality score of token $i$, instantiated in CDPruner as its relevance to the instruction.

Although these objectives encode diversity differently, both determine pairwise non-redundancy through the cosine geometry induced by $\bm S$. Max--min selection uses the cosine distance $D_{ij}=1-S_{ij}$, whereas determinantal selection constructs its diversity kernel from $\bm S$. This common reliance implicitly assumes that the visual features $\bm V$ produced by the encoder induce a similarity geometry suitable for diversity-based pruning. We therefore first investigate the geometry underlying visual-token similarities and how it affects diversity-based selection.



%% file: Section_4.tex
\section{Geometry Analysis and Cen-Prune}
\label{sec:analysis}

\begin{figure}
    \centering
    \includegraphics[width=\linewidth]{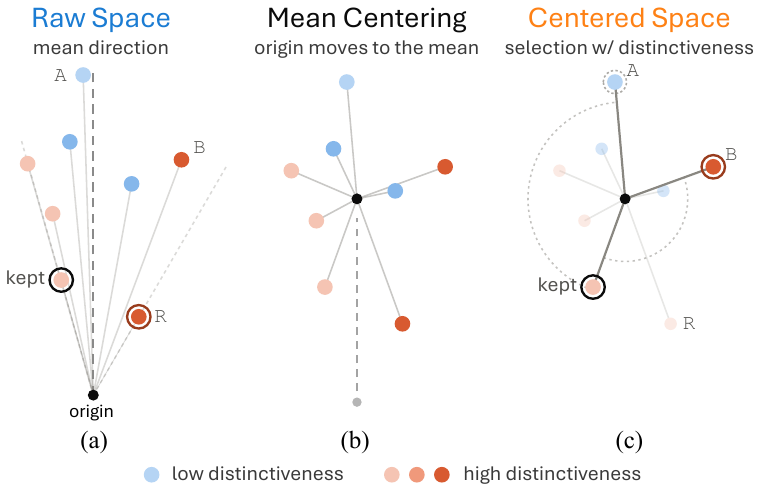}
    \caption{
    Raw and centered visual-token geometry.
    \textit{(a)} A shared mean direction limits angular contrast in raw space, leading max--min selection to \texttt{R}.
    \textit{(b)} Centering reveals diverse residual directions.
    \textit{(c)} Centered geometry alone may favor weakly distinctive \texttt{A}, whereas distinctiveness-aware selection favors \texttt{B}, which is both diverse and distinctive.
    Color indicates distinctiveness, measured by distance from the image-wise mean.
    }
    \label{fig:overview}
    \vspace{-5pt}
\end{figure}

We begin by examining the assumption raised above: whether the cosine-similarity matrix $\bm{S}$ induced by visual-token features provides a reliable geometry for diversity-based pruning. We analyze $\bm{S}$ computed from LLaVA-1.5-7B~\cite{llava/liu2023visual} visual tokens on 1,000 COCO val2017 images~\cite{COCO/lin2014microsoft}. Through this analysis, we uncover two key observations: (1) the raw similarity matrix is highly concentrated, limiting its ability to distinguish redundant from non-redundant tokens, and (2) centering token features alleviates this concentration but can suppress token-specific information in the original feature distribution. These findings motivate \textbf{Cen-Prune}, which combines subset diversity measured in the centered geometry with token distinctiveness captured by a raw-space distinctiveness score.




\begin{figure}
    \centering
    \includegraphics[width=0.85\linewidth]{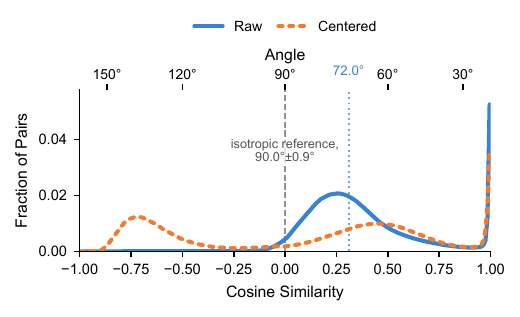}
    \vspace{-10pt}
    \caption{Cosine similarity over visual-token pairs. Frequencies are computed with 0.01-wide bins and smoothed for display. Raw features (blue) concentrate inside the acute range, with a median angle of $72^{\circ}$. Centered features (orange) shifts mass into the obtuse range, producing a negative lobe absent from the raw matrix.}
    \label{fig:fig2_full-cosine-similarity}
\end{figure}

\begin{table}[t]
    \centering
    \vspace{-5pt}
    \resizebox{0.95\linewidth}{!}{
    \begin{tabular}{lcccc}
        \toprule[1.2pt]
        \textbf{Selector} & \textbf{Mean Cos.} & \textbf{Obtuse (\%)} & \textbf{$<-0.5$ (\%)} & \textbf{$>0.99$ (\%)} \\
        \midrule
        All pairs     & 0.37 & 2.3 & 0.0 & 5.3 \\
        Random              & 0.37 & 2.3 & 0.0 & 5.3 \\
        MMDP       & 0.26 & 6.9 & 0.0 & 0.0 \\
        DPP   & 0.33 & 2.5 & 0.0 & 0.0 \\
        \bottomrule[1.2pt]
    \end{tabular}%
    }
    \caption{Geometry statistics of retained subsets selected by different diversity-based selectors (size = 32). Random is averaged over 10 samplings. MMDP with local search improves the max-min objective by only $2.6\%$.}
    \label{tab:selector_performance}
    \vspace{-10pt}
\end{table}

\subsection{Visual Similarity Is Highly Concentrated}

\paragraph{A Positive-Similarity Bias.}
We first examine the distribution of cosine similarities between distinct visual tokens in the raw similarity matrix $\bm S$. 
As a reference, two independently and uniformly sampled unit vectors in a 4,096-dimensional space are nearly orthogonal, with their angle concentrated around $90.0^\circ$ and a standard deviation of approximately $0.9^\circ$. However, as shown in Fig.~\ref{fig:fig2_full-cosine-similarity}, visual tokens deviate substantially from this reference. Their pairwise angles concentrate well below $90^\circ$, corresponding to predominantly positive cosine similarities, with a median angle of $72.0^\circ$ and a mean cosine similarity of $0.3663\pm0.0246$. Only $2.3\%$ of token pairs have negative cosine similarity, and none falls below $-0.5$ (Tab.~\ref{tab:selector_performance}). Meanwhile, $5.3\%$ of pairs are near-duplicates with cosine similarity above $0.99$.

A similar concentration is reported by \cite{WoldLike/lee2026more}, highlighting the geometric gap between visual and text tokens toward word-like image tokens. They report a mean cosine similarity of $0.3823\pm0.0018$ among LLaVA post-projector visual tokens, compared with $0.0378\pm0.0002$ among text tokens. Together, these results show that raw visual-token similarities occupy a narrow, predominantly positive range. As illustrated in Fig.~\ref{fig:overview}(a), this shared directional bias compresses the angular contrast available for diversity-based selection. The raw geometry clearly identifies near-duplicates, but provides limited contrast for distinguishing broader degrees of non-redundancy among the remaining tokens.


\paragraph{Selectors Only Deduplicate.}
One might still expect a strong selector to recover the few dissimilar token pairs hidden in the raw geometry. Tab.~\ref{tab:selector_performance} tests this possibility by selecting 32 tokens using MMDP from DivPrune~\cite{DivPrune/AlvarSAZ25} and DPP from CDPruner~\cite{CDPruner/zhang2026beyond}, with query relevance disabled for DPP. Random sampling closely reproduces the all-pairs statistics, confirming the broad homogeneity of the matrix.


MMDP eliminates near-duplicates, reducing the fraction of pairs with cosine similarity above $0.99$ from $5.3\%$ to $0.0\%$, and increases negative-cosine pairs from $2.3\%$ to $6.9\%$. Nevertheless, the selected subset still has a mean cosine similarity of $0.26$, with more than $93\%$ positive-cosine pairs. A single-swap local search changes these statistics only negligibly, indicating this limitation is not merely due to greedy search.

DPP exhibits a similar limitation. For two unit vectors, its Gram determinant is $1-\cos^2\theta$, making the objective insensitive to the sign of pairwise similarity. It therefore does not specifically favor negatively correlated pairs, yielding only $2.5\%$, close to random sampling. Across different objectives, these selectors effectively remove near-duplicates but fail to construct broadly dissimilar subsets. The primary bottleneck thus lies in the underlying geometry of $\bm S$.

\begin{figure}
    \centering
    \includegraphics[width=0.95\linewidth]{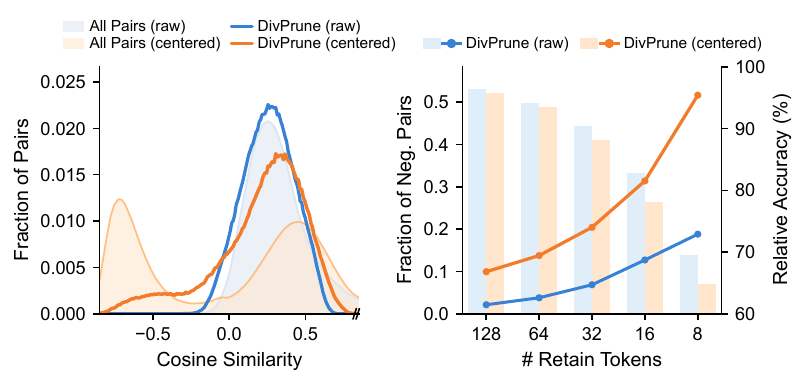}
    \caption{Comparison of selection on the raw/centered matrix. 
    \textit{(Left)} Cosine similarity distributions over selected subset ($n=32$) using DivPrune. Light shaded areas represent all pair reference. \textit{(Right)} As the retention budget tightens, the fraction of selected negative-cosine pairs rises under both matrices but far more steeply for centered selection (lines), while its relative accuracy stays below raw selection and the gap widens (bars).}
    \label{fig:fig3_comparison-on-both}
    \vspace{-5pt}
\end{figure}

\subsection{Mean Centering for Pure Diversification}

\paragraph{Centering Reveals Residual Diversity.}
The strong positive-similarity bias suggests removing the shared component across visual tokens. We therefore center each token feature as $\bar{\bm v}_i=\bm v_i-\bm\mu$, where $\bm\mu$ denotes the mean feature within an image. As illustrated in Fig.~\ref{fig:overview}(b), centering removes the common direction and exposes the residual variation among tokens. Consequently, pairwise cosine similarities spread substantially into the negative range (Fig.~\ref{fig:fig2_full-cosine-similarity}, orange). After centering, $32\%$ of all token pairs have cosine similarity below $-0.5$, with the distribution peaking near $-0.72$. Applying the same greedy MMDP procedure to the centered similarity matrix can therefore select tokens with strongly opposing residual directions, which are nearly absent from the raw geometry. In this sense, centering moves selection toward pure diversification: tokens are compared solely by their residual directions, allowing the selector to prioritize angular separation more directly.


\paragraph{Pure Diversification Is Not Enough.}
The broader angular range exposed by centering substantially changes the subsets produced by diversity-based selection. As shown in Fig.~\ref{fig:fig3_comparison-on-both}, the centered selector increasingly favors opposing residual directions as the token budget decreases. Under DivPrune, the fraction of negative-cosine pairs rises from $10.0\%$ at $n=128$ to $51.7\%$ at $n=8$, compared with an increase from $2.1\%$ to $18.8\%$ in the raw geometry. At $n=8$, the centered subset even exhibits a negative mean pairwise cosine similarity.

However, this stronger diversification does not translate into better pruning. At $n=32$, centered selection underperforms its raw counterpart on most benchmarks, and its relative accuracy remains lower across token budgets, with the gap widening as the budget becomes more restrictive (Fig.~\ref{fig:fig3_comparison-on-both}). This does not imply that negative-cosine pairs are undesirable. Rather, it shows that maximizing separation among residual directions alone is insufficient to determine which tokens should be retained.

\paragraph{Centering Leaves Out Distinctiveness.}
For each centered token feature $\bar{\bm v}_i=\bm v_i-\bm\mu$, selection relies on the cosine similarity: $\bar S_{ij}=
(\bar{\bm v}_i^\top \bar{\bm v}_j) /
(\|\bar{\bm v}_i\|\,\|\bar{\bm v}_j\|). 
$
Because this similarity normalizes each residual, it captures differences in residual direction but discards residual magnitude, $\|\bm v_i-\bm\mu\|$. This magnitude measures how strongly a token deviates from the common feature pattern shared across the image: tokens close to the mean largely follow this pattern and are more likely to be globally redundant, whereas tokens farther from the mean contain features less shared with the remaining tokens. Consequently, a weak residual can appear highly diverse solely because of its direction, while a token with a stronger and potentially more informative residual receives no additional preference, as illustrated in Fig.~\ref{fig:overview}(c). We view this residual magnitude as a token-level \emph{distinctiveness} signal, which we formalize as a distinctiveness score in Sec.~\ref{sec:method}. Although high distinctiveness does not necessarily imply semantic importance and may simply reflect an outlier, our empirical results show that highly distinctive tokens tend to capture semantically salient or globally non-redundant content, and that restoring this preference consistently improves centered selection. This suggests that centered directions and residual magnitudes provide complementary signals: the former captures subset diversity, while the latter provides a soft preference for individually distinctive tokens. Cen-Prune combines these signals to preserve the richer diversity of the centered geometry without discarding the useful token-wise preference retained in the original feature distribution.

\begin{figure}
    \centering
    \includegraphics[width=0.8\linewidth]{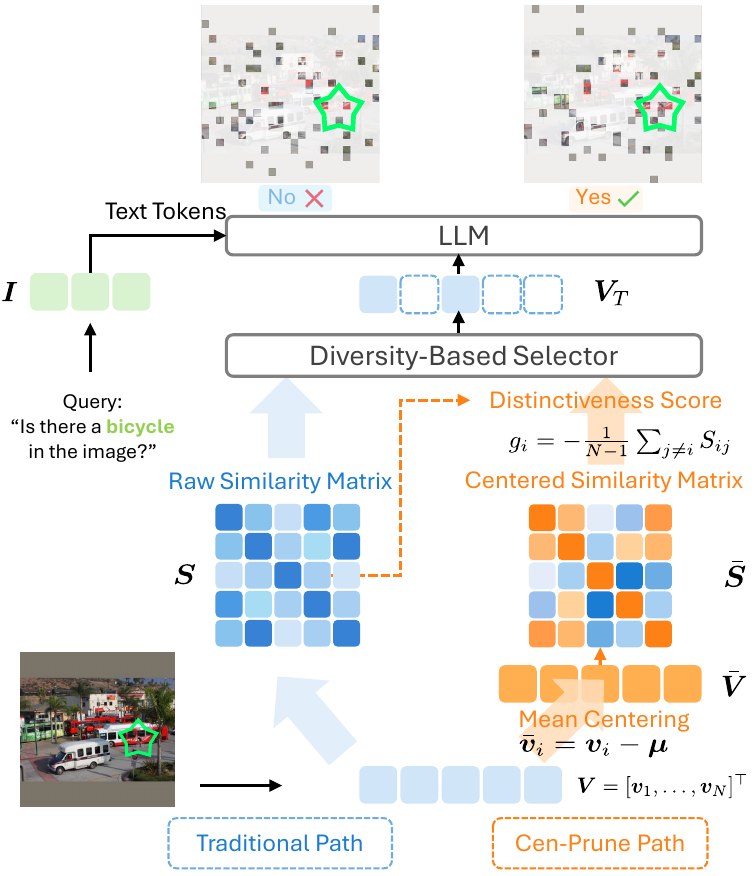}
    \caption{Overview of Cen-Prune. The traditional path feeds a diversity-based selector with the raw-feature similarity matrix ($\bm{S}$). Cen-Prune uses the shared mean direction twice: removed to build the selection matrix ($\bar{\bm{S}}$), and retained as a per-token distinctiveness score $g_i$ computed from $\bm{S}$. The selector is unchanged, yielding the retained tokens $\bm{V}_T$.}
    \label{fig:method}
    \vspace{-5pt}
\end{figure}

\subsection{Method: Cen-Prune} \label{sec:method}

Our analysis identifies two complementary roles of the shared component in visual-token features. Removing it exposes the angular contrast needed for subset diversity, while each token's relation to it provides a useful distinctiveness signal. Cen-Prune separates these roles: diversity is measured in the centered geometry, whereas distinctiveness is estimated from the uncentered geometry and incorporated as a soft selection preference. Figure~\ref{fig:method} summarizes this procedure.

\paragraph{Centered Geometry for Subset Diversity.}
Let $\bm X=[\boldsymbol{x}_1,\ldots,\boldsymbol{x}_N]^\top\in\mathbb{R}^{N\times D}$ 
denote the working token representation, which is either the original feature matrix $\bm V$ or its standardized variant described below. We first remove the image-wise feature mean:
\begin{equation}
    \bm\mu=\frac{1}{N}\sum_{j=1}^{N}\boldsymbol{x}_j,
    \qquad
    \bar{\boldsymbol{x}}_i=\boldsymbol{x}_i-\bm\mu.
\end{equation}
The centered similarity matrix is then constructed from the normalized residuals:
\begin{equation}
    \bar{S}_{ij}
    =
    \left\langle
    \hat{\bar{\boldsymbol{x}}}_i,
    \hat{\bar{\boldsymbol{x}}}_j
    \right\rangle,
    \qquad
    \hat{\bar{\boldsymbol{x}}}_i
    =
    \frac{\bar{\boldsymbol{x}}_i}{\|\bar{\boldsymbol{x}}_i\|_2}.
\end{equation}
By removing the shared component, $\bar{\bm S}$ exposes differences among residual directions and provides the angular contrast used to diversify the retained subset.

\paragraph{Token Distinctiveness from Uncentered Geometry.}
Centering is used only to construct the subset-diversity geometry. To retain the token-wise signal removed by this transformation, we measure how weakly each token aligns with the remaining tokens before centering. Let
$
    S_{ij}
    =
    \left\langle
    \hat{\boldsymbol{x}}_i,
    \hat{\boldsymbol{x}}_j
    \right\rangle
$
denote the uncentered cosine similarity. We define the distinctiveness score of token $i$ as its negated mean similarity to the other tokens:
\begin{equation}
    g_i
    =
    -\frac{1}{N-1}\sum_{j\neq i}S_{ij}.
    \label{eq:distinctiveness}
\end{equation}
This score has a direct geometric interpretation. Let
$\bm m=\frac{1}{N}\sum_j\hat{\boldsymbol{x}}_j$ denote the mean of the normalized uncentered features. Then,
\begin{equation}
    g_i
    =
    \frac{1-N\langle\hat{\boldsymbol{x}}_i,\bm m\rangle}{N-1},
\end{equation}
which is monotonically related to
$\|\hat{\bm{x}}_i-\bm m\|_2^2$.
Thus, a high $g_i$ identifies a token that lies far from the feature pattern shared across the image, whereas a low $g_i$ indicates that the token largely follows that pattern. We interpret this quantity as token-level \emph{distinctiveness}, rather than as a standalone measure of semantic importance, and use it only as a soft preference within diversity-based selection. The scores are min--max normalized within each image to obtain $\tilde g_i\in[0,1]$.

\paragraph{Integration with Diversity-Based Selectors.}
Cen-Prune preserves the optimization procedure of the underlying selector while replacing its raw geometry with $\bar{\bm S}$ and incorporating $\tilde{\bm g}$ as a token-wise quality signal. For the greedy MMDP selector used by DivPrune~\cite{DivPrune/AlvarSAZ25}, we construct the distance matrix $\bm{{D}}$ with entries $D_{ij} = 1 - \bar{S}_{ij}$.
After the selector's original initialization, each greedy step adds:
\begin{equation}
    i^\star
    =
    \arg\max_{i\notin T}
    \tilde g_i
    \min_{s\in T}D_{is},
    \label{eq:cen-mmdp}
\end{equation}
until $|T|=n$. A token is therefore preferred when it is both directionally separated from the retained set and individually distinctive in the original geometry. For the DPP-based selector used by CDPruner~\cite{CDPruner/zhang2026beyond}, we incorporate the same two signals into its kernel. Let $\tilde r_i$ denote the original instruction-relevance score. The corrected kernel is
\begin{equation}
    \bm L^{\mathrm{Cen}}
    =
    \operatorname{diag}
    (\tilde{\bm r}\odot\tilde{\bm g})
    \,
    \bar{\bm S}
    \,
    \operatorname{diag}
    (\tilde{\bm r}\odot\tilde{\bm g}),
    \label{eq:cen-dpp}
\end{equation}
where $\odot$ denotes element-wise multiplication. The original determinantal selection procedure is then applied to $\bm L^{\mathrm{Cen}}$. Thus, both selectors use centered directions to measure subset diversity and uncentered distinctiveness to determine which diverse tokens should receive preference.

%% file: Section_5.tex
\input{Tables/main_table_llava}

\section{Experiments}
\subsection{Experimental Setups}
\paragraph{Models and Baselines.}
We evaluate Cen-Prune across representative image and video LVLMs, including LLaVA-1.5~\cite{llava/liu2023visual}, LLaVA-NeXT~\cite{llavanext/liu2024llavanext}, LLaVA-Video~\cite{llava-video/ZhangWLLMLL25}, and Qwen2.5-VL~\cite{ qwen2.5/bai2025qwen25vltechnicalreport}. We compare against recent visual-token pruning methods: VisionZip~\cite{VisionZip/YangCTWL0J25}, HoloV~\cite{HoloV/abs-2510-02912}, VisPruner~\cite{VisPruner/ZhangCLZZCGSZ25}, DivPrune~\cite{DivPrune/AlvarSAZ25}, CDPruner~\cite{CDPruner/zhang2026beyond}, PruneSID~\cite{PruneSID/fang2026prune}, MMTok~\cite{MMTok/dong2026mmtok}, and ZOO-Prune~\cite{ZOO-Prune/Kim_2026_CVPR}. All methods are evaluated using their official implementations under a unified evaluation protocol.


\paragraph{Datasets.} Following prior work~\cite{ZOO-Prune/Kim_2026_CVPR,PruneSID/fang2026prune}, we evaluate image understanding on 11 benchmarks: GQA~\cite{GQA/hudson2019gqa}, MMBench~\cite{MMB/liu2024mmbench}, MME~\cite{MME/NEURIPS2025_d79a27cf}, POPE~\cite{POPE/li2023evaluating}, ScienceQA~\cite{SQA/lu2022learn}, VQAv2~\cite{VQAv2/goyal2017making}, TextVQA~\cite{TextVQA/singh2019towards}, OCRBench~\cite{OCR-B/liu2024ocrbench}, VizWiz~\cite{Vizwiz/gurari2018vizwiz}, SEED-Bench-Image~\cite{SEED/Li_2024_CVPR}, and MMMU~\cite{MMMU/yue2024mmmu}. We further evaluate LLaVA-Video-7B on four video-understanding benchmarks: MVBench~\cite{MVBench/li2024mvbench}, Video-MME~\cite{Video-MME/fu2025video}, the EgoSchema subset~\cite{Ego-Schema/NEURIPS2023_90ce332a}, and SEED-Bench-Video~\cite{SEED/Li_2024_CVPR}.


\paragraph{Implementation Details.} 
All experiments are conducted on NVIDIA A6000 GPUs and evaluated with \texttt{lmms-eval}. For LLaVA-NeXT-7B, which processes high-resolution images through up to five crops, we retain the same token proportion from each crop following VisionZip. Cen-Prune is training-free and has no tunable hyperparameters. Based on our geometry analysis, we apply the correction after the multimodal projector in LLaVA and before the visual merger in Qwen2.5-VL. Full details are provided in the Appendix.


\input{Tables/Qwen2.5}
\input{Tables/video}

\subsection{Main Results}
\paragraph{Image-Language Understanding.}
Tab.~\ref{tab:main_table_llava} shows that Cen-Prune provides robust performance improvements for existing diversity-based pruners across architectures and compression levels. The overall performance \textbf{Avg.} denotes the benchmark-wise retention ratio relative to the vanilla model, averaged over evaluated benchmarks. On \textbf{LLaVA-1.5-7B}, Cen-Prune with DivPrune and ZOO-Prune reaches 93.4\% and 93.6\% performance with only 32 retained tokens, matching the query-aware MMTok (93.4\%). Cen-Prune also improves CDPruner across all evaluated budgets. On \textbf{LLaVA-NeXT-7B}, it retains 96.1\% performance while removing 88.9\% of visual tokens. The gains extend to \textbf{Qwen2.5-VL-7B} with dynamic-resolution inputs: at 90\% token reduction, Cen-Prune retains 88.9\% performance, versus 85.1\% for DivPrune, 83.2\% for MMTok, and 79.5\% for CDPruner (Tab.~\ref{tab:qwen25_vl_single_column}). Additional LLaVA-1.5-13B results and other budgets are provided in the Appendix.


\paragraph{Video-Language Understanding.}
We further evaluate Cen-Prune on four video-understanding benchmarks using \textbf{LLaVA-Video-7B}. As shown in Tab.~\ref{tab:video_benchmarks_single_column}, Cen-Prune improves the overall performance of diversity-based pruning for video. At 90.5\% token reduction, Cen-Prune retains 92.2\% performance, exceeding DivPrune (87.9\%) and CDPruner (85.8\%).


\input{Tables/ablation}
\input{Tables/implicit_capture}
\input{Tables/efficiency}


\subsection{Ablation Studies and Further Analysis}
\paragraph{Component Ablation.}
Tab.~\ref{tab:combined_ablation_single_column} isolates the effects of mean centering and distinctiveness scoring. On \textbf{LLaVA-1.5-7B}, replacing the raw matrix with the centered matrix alone reduces performance, with the largest drop observed for DivPrune (greedy MMDP objective). CDPruner is less affected, consistent with the weaker sensitivity of its DPP objective to the sign of pairwise similarities. Distinctiveness scoring alone improves performance but does not address the concentrated geometry of raw features. On \textbf{Qwen2.5-VL-7B}, where DivPrune operates in a more concentrated embedding space, either centering or distinctiveness scoring alleviates degradation, while combining both yields the best performance.


\paragraph{Disentangling Diversity and Distinctiveness.}
To understand why the two components are complementary, Tab.~\ref{tab:departure_preference_analysis} analyzes the selected subsets in terms of residual geometry and token importance. Although vanilla DivPrune has no explicit importance term, its raw similarity matrix selects tokens with higher distinctiveness and greater \texttt{[CLS]} attention mass than the centered matrix (0.440 vs.\ 0.330 and 0.266 vs.\ 0.193, respectively). This indicates that the raw geometry implicitly favors important tokens, whereas centering weakens this preference while broadening the residual geometry. Adding ($\tilde{\bm g}$) to the centered selector restores the importance signal, achieving a \texttt{[CLS]} attention mass comparable to that of the raw selector with ($\tilde{\bm g}$) (0.311 vs.\ 0.312). At the same time, it preserves more diverse residual directions, with a lower mean cosine similarity (0.306 vs.\ 0.414) and a higher fraction of negative pairs (13.4\% vs.\ 7.1\%). Thus, Cen-Prune separates the importance signal from the diversity geometry rather than relying on their implicit coupling.

\paragraph{Efficiency Analysis.}
We finally measure computational efficiency on LLaVA-1.5-7B (Tab.~\ref{tab:efficiency_single_column}). We randomly sample 500 examples from each of TextVQA, POPE, MME, and GQA, and average each measurement over three runs. Relative to the full 576-token baseline, retaining 64 tokens with Cen-Prune yields a \(1.99\times\) prefill speedup, a \(1.67\times\) end-to-end speedup, and a 76.3\% FLOPs reduction. Its cost remains close to those of DivPrune and CDPruner, indicating that the geometry correction adds little overhead while providing substantially greater efficiency than ZOO-Prune.

%% file: Tables/main_table_llava.tex
\renewcommand{\multirowsetup}{\centering}
\newcommand{\categoryrule}{%
  \arrayrulecolor{gray!45}%
  \specialrule{0.4pt}{1.3pt}{1.3pt}%
  \arrayrulecolor{black}%
}
\newcommand{\doublethinrule}{%
  \specialrule{0.45pt}{1.5pt}{0.9pt}%
  \specialrule{0.45pt}{0pt}{1.5pt}%
}
\begin{table*}[!ht]
\centering

\renewcommand{\arraystretch}{0.88}
\setlength{\tabcolsep}{4.0pt}

\setlength{\aboverulesep}{0.2ex}
\setlength{\belowrulesep}{0.2ex}

\footnotesize
\centering
\scalebox{0.78}{
\begin{tabular}{l ccccccccccc c}
\toprule[1.2pt]

\textbf{Method}
& \textbf{GQA}
& \textbf{MMB}
& \textbf{MME}
& \textbf{POPE}
& \textbf{SQA}
& \textbf{VQA\textsuperscript{V2}}
& \textbf{VQA\textsuperscript{Text}}
& \textbf{OCR-B}
& \textbf{VizWiz}
& \textbf{SEED-I}
& \textbf{MMMU}
& \textbf{Avg.} \\
\midrule

\rowcolor{gray!10}
\textit{Upper Bound (576 Tokens)}
& 61.9 & 64.7 & 1511 & 85.9 & 69.6 & 78.5
& 58.2 & 31.3 & 54.4 & 66.2 & 36.3 & 100.0\% \\

\midrule

\rowcolor{gray!10}
\textbf{LLaVA-1.5-7B} & \multicolumn{12}{c}{\textit{Retain 128 Tokens} \textbf{\textcolor{mydarkgreen}{($\downarrow$ 77.8\%)}}} \\

VisionZip\confyear{CVPR25}
& 57.5 & 62.3 & 1439 & 83.1 & 68.6 & 75.6
& 56.9 & 30.0 & 54.4 & 61.5 & 37.8 & 97.0\% \\

HoloV\confyear{NeurIPS25}
& 57.4 & 63.0 & 1441 & 82.1 & 67.9 & 75.4
& 55.9 & 28.5 & 54.7 & 61.2 & 36.7 & 96.0\% \\

VisPruner\confyear{ICCV25}
& 58.4 & 62.5 & 1421 & 85.0 & 69.1 & 75.8
& 57.2 & 29.5 & 55.3 & 61.6 & 36.4 & 97.0\% \\

PruneSID\confyear{ICLR26}
& 58.2 & 62.2 & 1427 & 84.8 & 68.4 & 75.4
& 54.6 & 28.5 & 55.8 & 62.1 & 36.3 & 96.3\% \\

DivPrune\confyear{CVPR25}
& 59.3 & 62.4 & 1394 & 86.8 & 68.5 & 76.0
& 56.2 & 28.8 & 56.4 & 62.3 & 35.9 & 96.9\% \\

\rowcolor{mysoftgreen}
\quad \textbf{+ Ours}
& 59.3 & 62.5 & 1416 & 86.8 & 68.6 & 76.7
& 57.2 & 29.8 & 55.8 & 63.2 & 36.3 & 97.7\% \\

ZOO-Prune\confyear{CVPR26}
& 59.5 & 62.1 & 1387 & 86.8 & 68.5 & 76.4
& 56.8 & 29.1 & 56.0 & 63.0 & 35.8 & 97.1\% \\

\rowcolor{mysoftgreen}
\quad \textbf{+ Ours}
& 59.7 & 62.5 & 1445 & 86.5 & 68.9 & 76.7
& 57.0 & 29.5 & 55.5 & 63.5 & 36.9 & \textbf{98.0\%} \\

\categoryrule

MMTok\textsuperscript{*}\confyear{ICLR26}
& 59.3 & 61.9 & 1435 & 86.3 & 69.0 & 76.4
& 56.9 & 29.8 & 55.7 & 63.2 & 35.6 & 97.5\% \\

CDPruner\textsuperscript{*}\confyear{NeurIPS25}
& 59.6 & 62.4 & 1410 & 86.8 & 68.7 & 76.6
& 56.4 & 28.9 & 56.3 & 64.0 & 35.3 & 97.3\% \\

\rowcolor{mysoftgreen}
\quad \textbf{+ Ours}
& 60.2 & 63.1 & 1442 & 86.7 & 68.2 & 76.9
& 57.0 & 30.0 & 55.4 & 63.9 & 36.1 & \textbf{98.1\%} \\

\midrule

\rowcolor{gray!10}
\textbf{LLaVA-1.5-7B} & \multicolumn{12}{c}{\textit{Retain 64 Tokens} \textbf{\textcolor{mydarkgreen}{($\downarrow$ 88.9\%)}}} \\


VisionZip\confyear{CVPR25}
& 55.1 & 60.1 & 1370 & 77.1 & 69.0 & 72.4
& 55.4 & 28.3 & 54.8 & 57.8 & 36.4 & 93.4\% \\

HoloV\confyear{NeurIPS25}
& 55.1 & 60.2 & 1366 & 76.9 & 68.8 & 72.7
& 55.1 & 27.1 & 55.8 & 58.2 & 36.1 & 93.1\% \\

VisPruner\confyear{ICCV25}
& 55.6 & 59.6 & 1357 & 80.4 & 68.5 & 72.6
& 55.9 & 29.2 & 56.6 & 58.0 & 35.7 & 94.1\% \\

PruneSID\confyear{ICLR26}
& 57.7 & 60.1 & 1401 & 84.5 & 68.8 & 73.8
& 54.7 & 27.2 & 56.5 & 60.2 & 37.2 & 95.3\% \\

DivPrune\confyear{CVPR25}
& 57.8 & 59.5 & 1351 & 85.6 & 68.0 & 74.1
& 54.7 & 27.4 & 57.4 & 60.3 & 35.7 & 94.8\% \\

\rowcolor{mysoftgreen}
\quad \textbf{+ Ours}
& 58.4 & 60.5 & 1370 & 85.3 & 68.5 & 75.3
& 56.6 & 28.5 & 57.4 & 60.8 & 36.2 & 96.2\% \\

ZOO-Prune\confyear{CVPR26}
& 58.4 & 60.1 & 1375 & 85.7 & 68.2 & 74.9
& 55.6 & 27.6 & 57.3 & 61.0 & 35.6 & 95.5\% \\

\rowcolor{mysoftgreen}
\quad \textbf{+ Ours}
& 58.2 & 60.7 & 1390 & 85.3 & 67.8 & 75.3
& 56.2 & 28.4 & 56.9 & 61.7 & 36.1 & \textbf{96.2\%} \\

\categoryrule

MMTok\textsuperscript{*}\confyear{ICLR26}
& 58.2 & 60.4 & 1404 & 85.8 & 68.7 & 75.2
& 55.7 & 28.6 & 57.2 & 61.5 & 36.0 & 96.3\% \\

CDPruner\textsuperscript{*}\confyear{NeurIPS25}
& 58.8 & 61.3 & 1400 & 87.0 & 68.5 & 75.4
& 55.4 & 28.0 & 57.0 & 62.6 & 35.4 & 96.3\% \\

\rowcolor{mysoftgreen}
\quad \textbf{+ Ours}
& 59.0 & 61.9 & 1414 & 86.1 & 68.8 & 76.1
& 56.3 & 28.1 & 56.9 & 62.7 & 36.3 & \textbf{97.0\%} \\

\midrule

\rowcolor{gray!10}
\textbf{LLaVA-1.5-7B} & \multicolumn{12}{c}{\textit{Retain 32 Tokens} \textbf{\textcolor{mydarkgreen}{($\downarrow$ 94.4\%)}}} \\


VisionZip\confyear{CVPR25}
& 51.8 & 57.3 & 1243 & 69.1 & 68.8 & 67.1
& 53.2 & 25.4 & 55.7 & 53.2 & 35.1 & 88.3\% \\

HoloV\confyear{NeurIPS25}
& 52.8 & 59.0 & 1262 & 70.5 & 68.9 & 68.7
& 53.7 & 25.9 & 56.3 & 55.0 & 36.0 & 89.9\% \\

VisPruner\confyear{ICCV25}
& 51.6 & 56.9 & 1261 & 73.1 & 68.1 & 67.6
& 53.6 & 25.8 & 55.8 & 53.7 & 36.2 & 89.3\% \\

PruneSID\confyear{ICLR26}
& 54.7 & 56.0 & 1297 & 79.9 & 67.2 & 70.5
& 52.4 & 24.9 & 56.7 & 56.9 & 35.3 & 90.7\% \\

DivPrune\confyear{CVPR25}
& 55.0 & 58.3 & 1306 & 81.7 & 68.4 & 71.1
& 53.1 & 25.5 & 57.0 & 57.2 & 35.2 & 91.8\% \\

\rowcolor{mysoftgreen}
\quad \textbf{+ Ours}
& 56.0 & 59.5 & 1335 & 82.8 & 68.6 & 72.9
& 55.3 & 26.2 & 57.6 & 58.6 & 34.9 & 93.4\% \\

ZOO-Prune\confyear{CVPR26}
& 56.0 & 57.7 & 1313 & 83.0 & 69.1 & 72.3
& 53.8 & 26.1 & 57.2 & 57.9 & 34.9 & 92.6\% \\

\rowcolor{mysoftgreen}
\quad \textbf{+ Ours}
& 56.2 & 59.8 & 1354 & 83.1 & 69.2 & 73.1
& 55.0 & 26.2 & 57.1 & 58.6 & 34.9 & \textbf{93.6\%} \\

\categoryrule

MMTok\textsuperscript{*}\confyear{ICLR26}
& 56.2 & 58.8 & 1358 & 82.9 & 69.0 & 73.1
& 53.7 & 26.4 & 57.2 & 59.6 & 34.8 & 93.4\% \\

CDPruner\textsuperscript{*}\confyear{NeurIPS25}
& 57.2 & 60.1 & 1359 & 87.2 & 69.0 & 73.7
& 52.9 & 24.4 & 56.5 & 61.0 & 35.2 & 93.8\% \\

\rowcolor{mysoftgreen}
\quad \textbf{+ Ours}
& 57.5 & 60.6 & 1372 & 86.1 & 68.2 & 74.8
& 55.2 & 27.1 & 56.8 & 61.1 & 36.1 & \textbf{95.4\%} \\

\doublethinrule

\rowcolor{gray!10}
\textit{Upper Bound (2880 Tokens)}
& 64.2 & 67.9 & 1520 & 86.4 & 70.2 & 80.1
& 61.3 & 51.3 & 59.7 & 69.6 & 37.2 & 100.0\% \\

\midrule

\rowcolor{gray!10}
\textbf{LLaVA-NeXT-7B} & \multicolumn{12}{c}{\textit{Retain 320 Tokens} \textbf{\textcolor{mydarkgreen}{($\downarrow$ 88.9\%)}}} \\


VisionZip\confyear{CVPR25}
& 58.9 & 63.4 & 1401 & 82.1 & 67.3 & 76.2
& 58.8 & 39.2 & 59.9 & 63.5 & 36.7 & 93.2\% \\

PruneSID\confyear{ICLR26}
& 60.6 & 63.0 & 1468 & 85.0 & 67.8 & 77.1
& 54.3 & 32.4 & 58.7 & 65.8 & 36.9 & 92.6\% \\

VisPruner\confyear{ICCV25}
& 58.6 & 63.7 & 1404 & 81.1 & 67.9 & 75.8
& 58.6 & 37.5 & 61.1 & 62.8 & 35.7 & 92.7\% \\

DivPrune\confyear{CVPR25}
& 61.1 & 63.7 & 1375 & 84.6 & 67.8 & 77.1
& 56.3 & 34.2 & 59.7 & 65.7 & 36.7 & 92.9\% \\

\rowcolor{mysoftgreen}
\quad \textbf{+ Ours}
& 60.9 & 64.6 & 1480 & 85.1 & 68.0 & 78.3
& 57.6 & 39.3 & 59.2 & 66.4 & 36.2 & \textbf{94.8\%} \\

ZOO-Prune\confyear{CVPR26}
& 61.0 & 64.1 & 1413 & 85.3 & 67.4 & 77.4
& 57.1 & 36.7 & 58.9 & 65.8 & 37.2 & 93.8\% \\

\rowcolor{mysoftgreen}
\quad \textbf{+ Ours}
& 60.8 & 64.6 & 1443 & 85.4 & 67.8 & 78.1
& 58.0 & 39.6 & 58.9 & 66.2 & 36.3 & 94.6\% \\
 
\categoryrule

MMTok\textsuperscript{*}\confyear{ICLR26}
& 61.1 & 64.1 & 1471 & 85.8 & 67.6 & 77.7
& 56.9 & 34.6 & 59.1 & 66.2 & 38.0 & 94.1\% \\

CDPruner\textsuperscript{*}\confyear{NeurIPS25}
& 61.6 & 64.2 & 1485 & 87.3 & 66.8 & 78.5
& 57.0 & 38.8 & 58.4 & 67.3 & 35.8 & 94.7\% \\

\rowcolor{mysoftgreen}
\quad \textbf{+ Ours}
& 61.4 & 65.3 & 1490 & 86.8 & 67.2 & 79.0
& 58.9 & 43.3 & 58.8 & 67.7 & 35.9 & \textbf{96.1\%} \\

\bottomrule[1.2pt]
\end{tabular}}
\caption{Performance comparison on LLaVA-1.5-7B and LLaVA-NeXT-7B across 11 multimodal benchmarks. Methods marked with \textsuperscript{*} are query-aware. The \textbf{best} average result within each category is highlighted in bold.}
\label{tab:main_table_llava}
     \vspace{-5pt}
\end{table*}

%% file: Tables/Qwen2.5.tex

\begin{table}[!ht]
\centering

\setlength{\tabcolsep}{2.5pt} 
\renewcommand{\arraystretch}{0.98} 

\footnotesize

\setlength{\aboverulesep}{0.2ex}
\setlength{\belowrulesep}{0.2ex}

\scalebox{0.7}{
\begin{tabular}{l ccccccc c}
\toprule[1.2pt]
\textbf{Method} & \textbf{GQA} & \textbf{MMB} & \textbf{MME} & \textbf{POPE} & \textbf{SQA} & \textbf{VQA\textsuperscript{Text}} & \textbf{OCR-B} & \textbf{Avg.} \\
\midrule
\rowcolor{gray!10} \textit{Upper Bound (100\%)} & 60.3 & 83.2 & 2322 & 86.2 & 87.4 & 77.5 & 83.8 & 100.0\% \\
\midrule

\rowcolor{gray!10} \textbf{Qwen2.5-VL-7B} & \multicolumn{7}{c}{\textit{Retain 20\% Tokens}} & \\
DivPrune\confyear{CVPR25} & 58.5 & 78.3 & 2158 & 83.4 & 84.6 & 70.5 & 64.1 & 92.1\% \\
\rowcolor{mysoftgreen} \quad \textbf{+ Ours} & 58.2 & 79.6 & 2274 & 84.0 & 84.2 & 73.7 & 68.9 & \textbf{94.5\%} \\
CDPruner\confyear{NeurIPS25} & 56.2 & 78.4 & 2110 & 80.2 & 84.2 & 63.6 & 50.5 & 87.1\% \\
\rowcolor{mysoftgreen} \quad \textbf{+ Ours} & 57.4 & 79.5 & 2227 & 81.2 & 84.5 & 70.0 & 59.4 & 91.3\% \\
MMTok\confyear{ICLR26} & 57.2 & 79.6 & 2216 & 82.3 & 83.7 & 71.6 & 59.3 & 91.5\% \\
\midrule

\rowcolor{gray!10} \textbf{Qwen2.5-VL-7B} & \multicolumn{7}{c}{\textit{Retain 10\% Tokens}} & \\
DivPrune\confyear{CVPR25} & 55.1 & 74.8 & 2035 & 79.0 & 82.1 & 65.0 & 48.1 & 85.1\% \\
\rowcolor{mysoftgreen} \quad \textbf{+ Ours} & 55.6 & 75.9 & 2156 & 82.0 & 82.4 & 69.2 & 56.7 & \textbf{88.9\%} \\
CDPruner\confyear{NeurIPS25} & 53.0 & 75.9 & 1925 & 74.4 & 83.0 & 56.5 & 33.9 & 79.5\% \\
\rowcolor{mysoftgreen} \quad \textbf{+ Ours} & 54.0 & 74.5 & 2076 & 76.9 & 83.0 & 63.4 & 47.4 & 84.4\% \\
MMTok\confyear{ICLR26} & 53.9 & 74.7 & 1992 & 77.1 & 81.5 & 64.6 & 43.2 & 83.2\% \\
\midrule

\rowcolor{gray!10} \textbf{Qwen2.5-VL-7B} & \multicolumn{7}{c}{\textit{Retain 5\% Tokens}} & \\
DivPrune\confyear{CVPR25} & 50.4 & 67.5 & 1853 & 69.5 & 78.8 & 57.3 & 33.3 & 75.6\% \\
\rowcolor{mysoftgreen} \quad \textbf{+ Ours} & 51.7 & 70.1 & 1944 & 75.0 & 79.2 & 63.4 & 42.4 & \textbf{80.5\%} \\
CDPruner\confyear{NeurIPS25} & 48.5 & 70.1 & 1724 & 64.7 & 81.2 & 50.4 & 20.5 & 70.9\% \\
\rowcolor{mysoftgreen} \quad \textbf{+ Ours} & 48.6 & 66.8 & 1773 & 67.5 & 80.8 & 56.7 & 30.9 & 74.0\% \\
MMTok\confyear{ICLR26} & 48.4 & 68.3 & 1794 & 68.4 & 78.9 & 57.9 & 29.6 & 74.2\% \\
\bottomrule[1.2pt]
\end{tabular}%
}
\vspace{-5pt}
\caption{Performance comparison on Qwen2.5-VL-7B.}
\label{tab:qwen25_vl_single_column}
\vspace{-10pt}
\end{table}

%% file: Tables/video.tex
\begin{table}[!ht]
\centering

\setlength{\aboverulesep}{0.2ex}
\setlength{\belowrulesep}{0.2ex}

\setlength{\tabcolsep}{2.0pt} 
\renewcommand{\arraystretch}{0.98} 

\footnotesize
\scalebox{0.73}{
\begin{tabular}{l  cccc  c}
\toprule[1.2pt]
\textbf{Method} & \textbf{MVBench} & \textbf{MME-V} & \textbf{EgoSchema} & \textbf{SEED-V} & \textbf{Avg.} \\
\midrule
\rowcolor{gray!10} \textit{Upper Bound $64 \times 169$ Tokens} & 60.3 & 64.3 & 59.2 & 58.1 & 100.0\% \\
\midrule

\rowcolor{gray!10} \textbf{LLaVA-Video-7B} & \multicolumn{4}{ c}{\textit{Retain 64 $\times$ 64 Tokens} \textbf{\textcolor{mydarkgreen}{($\downarrow$ 62.1\%)}}} & \multicolumn{1}{c}{} \\ 
DivPrune\confyear{CVPR25} & 54.9 & 61.6 & 59.2 & 56.3 & 95.9\% \\
\rowcolor{mysoftgreen} \quad + \textbf{Ours} & 56.4 & 62.1 & 60.6 & 58.0 & \textbf{98.1\%} \\
CDPruner\confyear{NeurIPS25} & 54.9 & 60.4 & 58.2 & 56.3 & 95.1\% \\
\rowcolor{mysoftgreen} \quad + \textbf{Ours} & 55.6 & 60.3 & 60.8 & 56.6 & 96.6\% \\
\midrule

\rowcolor{gray!10} \textbf{LLaVA-Video-7B} & \multicolumn{4}{ c}{\textit{Retain 64 $\times$ 32 Tokens} \textbf{\textcolor{mydarkgreen}{($\downarrow$ 81.1\%)}}} & \multicolumn{1}{c}{} \\ 
DivPrune\confyear{CVPR25} & 54.1 & 58.9 & 57.2 & 55.2 & 93.2\% \\
\rowcolor{mysoftgreen} \quad + \textbf{Ours} & 54.8 & 60.2 & 59.2 & 56.2 & \textbf{95.3\%} \\
CDPruner\confyear{NeurIPS25} & 53.4 & 57.5 & 56.2 & 52.9 & 91.0\% \\
\rowcolor{mysoftgreen} \quad + \textbf{Ours} & 53.4 & 57.6 & 57.6 & 53.0 & 91.7\% \\
\midrule

\rowcolor{gray!10} \textbf{LLaVA-Video-7B} & \multicolumn{4}{c}{\textit{Retain 64 $\times$ 16 Tokens} \textbf{\textcolor{mydarkgreen}{($\downarrow$ 90.5\%)}}} & \multicolumn{1}{c}{} \\ 
DivPrune\confyear{CVPR25} & 51.9 & 55.9 & 53.2 & 51.4 & 87.9\% \\
\rowcolor{mysoftgreen} \quad + \textbf{Ours} & 52.9 & 58.1 & 58.0 & 53.7 & \textbf{92.2\%} \\
CDPruner\confyear{NeurIPS25} & 50.9 & 54.4 & 53.6 & 48.5 & 85.8\% \\
\rowcolor{mysoftgreen} \quad + \textbf{Ours} & 50.3 & 54.8 & 55.4 & 48.4 & 86.4\% \\
\bottomrule[1.2pt]
\end{tabular}}
\vspace{-5pt}
\caption{Performance comparison on LLaVA-Video-7B.}
\label{tab:video_benchmarks_single_column}
\vspace{-4pt}
\end{table}

%% file: Tables/ablation.tex
\newcommand{\doublethinruleforablation}{%
  \specialrule{0.40pt}{1.5pt}{0.9pt}%
  \specialrule{0.40pt}{0pt}{1.5pt}%
}
\begin{table}[!t]
\centering
\scriptsize 
\setlength{\tabcolsep}{7.0pt} 
\renewcommand{\arraystretch}{0.98} 
\footnotesize

\setlength{\aboverulesep}{0.2ex}
\setlength{\belowrulesep}{0.2ex}

\scalebox{0.75}{
\begin{tabular}{l cccc c}
\toprule[1.2pt]

\textbf{Method} & \textbf{GQA} & \textbf{MME} & \textbf{POPE} & \textbf{VQA\textsuperscript{Text}} & \textbf{Avg.} \\
\midrule

\rowcolor{gray!10} \textbf{LLaVA-1.5-7B} & \multicolumn{4}{c}{\textit{Retain 32 Tokens} \textbf{\textcolor{mydarkgreen}{($\downarrow$ 94.4\%)}}} & \multicolumn{1}{c}{} \\ 
DivPrune\confyear{CVPR25} & 55.0 & 1306 & \underline{81.7} & 53.1 & 90.4\% \\
\quad + centering & 54.5 & 1263 & 80.6 & 50.9 & 88.2\% \\
\quad + distinctiveness & \underline{55.4} & \underline{1329} & 81.4 & \underline{55.0} & \underline{91.7\%} \\
\rowcolor{mysoftgreen} Cen-Prune (Ours) & \textbf{56.0} & \textbf{1335} & \textbf{82.8} & \textbf{55.3} & \textbf{92.6\%} \\
\midrule

CDPruner\confyear{NeurIPS25} & 57.2 & \underline{1359} & 87.2 & 52.9 & 93.7\% \\
\quad + centering & 56.9 & 1354 & \textbf{87.4} & 53.1 & 93.6\% \\
\quad + distinctiveness & \underline{57.2} & 1359 & 85.4 & \underline{54.9} & \underline{94.0\%} \\
\rowcolor{mysoftgreen} Cen-Prune (Ours) & \textbf{57.5} & \textbf{1372} & \underline{86.1} & \textbf{55.2} & \textbf{94.7\%} \\

\doublethinruleforablation

\rowcolor{gray!10} \textbf{Qwen2.5-VL-7B} & \multicolumn{4}{c}{\textit{Retain 10\% Tokens}} & \multicolumn{1}{c}{} \\ 
DivPrune\confyear{CVPR25} & \underline{55.1} & \underline{2035} & 79.0 & \underline{65.0} & \underline{88.6\%} \\
DivPrune\confyear{CVPR25}\textdagger & 54.6 & 1951 & 78.3 & 61.3 & 86.1\% \\
\quad + centering & 54.1 & 2024 & \underline{79.5} & 61.3 & 87.0\% \\
\quad + distinctiveness & 54.4 & 1997 & 78.3 & 63.1 & 87.1\% \\
\rowcolor{mysoftgreen} Cen-Prune (Ours) & \textbf{55.6} & \textbf{2156} & \textbf{82.0} & \textbf{69.2} & \textbf{92.4\%} \\

\bottomrule[1.2pt]
\end{tabular}
}
\vspace{-5pt}
    \caption{Ablation study on LLaVA-1.5-7B and Qwen2.5-VL-7B. \textbf{Best} and \underline{second-best} results within each group are highlighted. \dag~denotes pre-projector token pruning.}
    \label{tab:combined_ablation_single_column}
\vspace{-10pt}
\end{table}

%% file: Tables/implicit_capture.tex


\begin{table}[t]
\centering
\vspace{-1pt}

\footnotesize                      
\setlength{\tabcolsep}{2.0pt}        

\renewcommand{\arraystretch}{0.98}   
\resizebox{\linewidth}{!}{%
\begin{tabular}{l cccc}
\toprule[1.2pt]
  \textbf{Method} &
  \textbf{Res. Mean Cos.} &
  \textbf{Res. Neg. (\%)} &
  \textbf{Mean Distinct.} &
  \textbf{\texttt{[CLS]} Attn.}
 \\
\midrule

DivPrune (raw)       & 0.330 & 12.0 & 0.440 & 0.266 \\
DivPrune (raw) w/ $\tilde{\bm{g}}$  & 0.414 &  7.1 & 0.533 & 0.312 \\

DivPrune (centered)      & 0.203 & 20.4 & 0.330 & 0.193 \\
DivPrune (centered) w/ $\tilde{\bm{g}}$ & 0.306 & 13.4 & 0.499 & 0.311 \\

\bottomrule[1.2pt]
\end{tabular}}
\vspace{-5pt}
\caption{Disentangling the effects of the similarity matrix and the distinctiveness preference on COCO val2017.}
\label{tab:departure_preference_analysis}
\vspace{-5pt}
\end{table}

%% file: Tables/efficiency.tex
\begin{table}[t]
    \centering
    \vspace{-1pt}
    
\setlength{\aboverulesep}{0.2ex}
\setlength{\belowrulesep}{0.2ex}

\setlength{\tabcolsep}{4.0pt} 
\renewcommand{\arraystretch}{0.98} 

\footnotesize
\scalebox{0.73}{
\begin{tabular}{l  cccc}
\toprule[1.2pt]
\textbf{Method} & \textbf{\#Tokens} & \textbf{Prefill (ms)} & \textbf{Latency (ms)} & \textbf{FLOPs (G)} \\
\midrule
\rowcolor{gray!10} \textbf{LLaVA-1.5-7B} & 576 & 154.77 & 200.27 & 4431.21 \\
DivPrune\confyear{CVPR25} & 64 & 77.68 & 119.53 & 1048.38 \\
\rowcolor{mysoftgreen} \quad + \textbf{Ours} & 64 & 77.82 & 119.85 & 1048.38 \\
CDPruner\confyear{NeurIPS25} & 64 & 90.92 & 133.54 & 1056.75 \\
\rowcolor{mysoftgreen} \quad + \textbf{Ours} & 64 & 91.85 & 134.63 & 1056.75 \\
ZOO-Prune\confyear{CVPR26} & 64 & 112.08 &154.51 & 2595.93 \\
\rowcolor{mysoftgreen} \quad + \textbf{Ours} & 64 & 112.32 & 154.77 & 2595.93 \\
\bottomrule[1.2pt]
\end{tabular}}
    \vspace{-5pt}
    \caption{Efficiency comparison on LLaVA-1.5-7B.}
    \label{tab:efficiency_single_column}
    \vspace{-5pt}
\end{table}

%% file: Section_6.tex
\section{Conclusion}
We examine the visual-token geometry underlying diversity-based pruning and find that the raw cosine-similarity matrix is strongly concentrated in the positive range. This provides limited resolution for residual diversity while implicitly favoring globally distinctive tokens, thereby entangling pairwise diversity with token-wise importance. Centering resolves residual diversity but weakens this useful preference. Accordingly, we introduce Cen-Prune, which measures subset diversity in the centered geometry while retaining raw-space distinctiveness for token selection. Cen-Prune is training-free, preserves the underlying selector, and adds little computational overhead. Experiments across LVLM architectures and image and video benchmarks demonstrate robust improvements. More broadly, our results suggest that diversity-based pruning benefits from modeling subset diversity and token-wise distinctiveness in complementary geometries.

%% file: Appendix.tex
\clearpage

\setcounter{table}{0}
\setcounter{figure}{0}

\renewcommand{\thetable}{A\arabic{table}}
\renewcommand{\thefigure}{A\arabic{figure}}

\appendix

\twocolumn[
\begin{center}
{\LARGE\bfseries Centering before Pruning: Lightweight Geometry Correction
for Diversity-Based Visual Token Pruning in LVLMs\par}
\vspace{0.7em}
{\Large Supplementary Material\par}
\vspace{1.5em}
\end{center}
]

\section*{Appendix}
This supplementary material provides additional empirical analyses and benchmark evaluation results, organized as follows:
\begin{itemize}
\item In Sec.~\ref{app:empirical_analysis}, we provide the details of our empirical analysis.
\item In Sec.~\ref{app:implementation_details}, we present additional implementation details for the benchmark evaluations. 
\item In Sec.~\ref{app:additional_results}, we also report further results, including performance comparisons on larger-scale LVLMs, additional ablation studies and further efficiency analysis.
\item In Sec.~\ref{app:visualization}, we present the qualitative visualizations of DivPrune and our Cen-Prune method.
\end{itemize}
\input{appendix_A}

\section{Details of Experimental Setup} \label{app:implementation_details}

\subsection{Model Settings}
We evaluate our visual token pruning method on representative open-source LVLMs covering image-only, high-resolution image, and video understanding settings. Following prior visual-token pruning studies, we keep the original model weights frozen and apply token pruning only during inference, so that any performance change reflects the effect of reducing visual tokens rather than additional training or fine-tuning.

\paragraph{LLaVA-1.5~\citeapp{llava/liu2023visual}.} We use LLaVA-1.5 as the standard image-understanding backbone. LLaVA-1.5 follows the widely used LLaVA architecture, consisting of a CLIP ViT-L/336px vision encoder~\citeapp{CLIP/radford2021learning}, a two-layer MLP vision-language projector, and a Vicuna language model. Compared with the original LLaVA, LLaVA-1.5 improves the visual instruction tuning recipe by using stronger image resolution, an MLP connector, and additional academic-task-oriented VQA data. In our experiments, this model serves as the main low-resolution image baseline, where each image is encoded into a fixed number of visual tokens. This setting is commonly adopted in visual token pruning literature because it provides a controlled testbed for evaluating the accuracy-efficiency trade-off under different retained-token budgets.

\paragraph{LLaVA-1.6 (LLaVA-NeXT)~\citeapp{llavanext/liu2024llavanext}.} We further evaluate on LLaVA-1.6, also known as LLaVA-NeXT, to test pruning under high-resolution image inputs. LLaVA-NeXT extends LLaVA-style visual instruction tuning with an AnyRes image representation, where an image can be split into multiple visual grids before being encoded. This design preserves more fine-grained visual details than the fixed-resolution setting, but also substantially increases the number of visual tokens passed to the language model. Therefore, LLaVA-NeXT provides a more challenging and practically important setting for visual token pruning: the model has stronger visual perception ability, while its longer visual prefix leads to higher prefilling cost and memory usage.

\paragraph{LLaVA-Video~\citeapp{llava-video/ZhangWLLMLL25}} For video understanding, we use \texttt{lmms-lab/LLaVA-Video-7B-Qwen2}. This model is built on the LLaVA-Video framework, which adapts LLaVA-style multimodal instruction tuning to video inputs. The released checkpoint uses a SigLIP/SO400M vision encoder~\citeapp{Siglip/zhai2023sigmoid} with a Qwen2~\citeapp{Qwen2/yang2024qwen2technicalreport} language backbone and is initialized from the LLaVA-OneVision~\citeapp{LLaVA-OneVision/0080ZGZ00ZZL0L25} image model before being trained on a mixture of single-image, multi-image, and video instruction data. In our experiments, LLaVA-Video-7B-Qwen2 is used to evaluate whether visual token pruning remains effective when the visual input consists of multiple sampled video frames rather than a single image. This setting is important because video models naturally introduce many more visual tokens, making inference cost more severe.

\paragraph{Qwen2.5-VL~\citeapp{qwen2.5/bai2025qwen25vltechnicalreport}.}
We also evaluate Qwen2.5-VL-7B as a stronger recent MLLM backbone. Qwen2.5-VL introduces a native dynamic-resolution vision encoder, window attention in the vision transformer, and improved spatial-temporal modeling for images and videos. Unlike LLaVA-style models that typically rely on a fixed CLIP-like image encoder, Qwen2.5-VL dynamically converts visual inputs of different sizes into variable-length visual tokens. This makes it a useful test case for evaluating whether our pruning strategy generalizes beyond the LLaVA family. In this work, we report results on the \texttt{Qwen2.5-VL-7B-Instruct}, aligning its model scale with the other evaluated backbones while isolating the effect of architectural differences.

\subsection{Datasets}
\subsubsection{Image Benchmarks.}
Following prior studies~\citeapp{ZOO-Prune/Kim_2026_CVPR,PruneSID/fang2026prune} on efficient multimodal large language models, we evaluate image understanding on 11 widely used benchmarks covering general VQA, compositional reasoning, OCR-centric perception, hallucination evaluation, scientific reasoning, real-world assistive VQA, and expert-level multimodal understanding. We also list the specific dataset partitions we used with \texttt{lmms-eval}\footnote{https://github.com/EvolvingLMMs-Lab/lmms-eval} as below.
\begin{itemize}
\renewcommand\labelitemi{$\bullet$}
\item \textbf{GQA}~\citeapp{GQA/hudson2019gqa} evaluates real-world visual reasoning and compositional question answering. It requires models to understand objects, attributes, relations, and scene structures, making it suitable for testing fine-grained visual reasoning.

\item \textbf{MMBench}~\citeapp{MMB/liu2024mmbench} is a multiple-choice benchmark for evaluating fine-grained multimodal abilities. It covers broad perception and reasoning dimensions, providing an objective evaluation protocol for LVLMs. We use \texttt{mmbench\_en\_dev} for evaluations.

\item \textbf{MME}~\citeapp{MME/NEURIPS2025_d79a27cf} provides a comprehensive evaluation of LVLMs from both perception and cognition perspectives. Its perception tasks include object existence, color, position, counting, and OCR, while its cognition tasks include commonsense reasoning, numerical calculation, text translation, and code reasoning.

\item \textbf{POPE}~\citeapp{POPE/li2023evaluating} evaluates object hallucination in LVLMs through binary object-presence questions. It tests whether a model incorrectly asserts the existence of objects that are absent from the image.

\item \textbf{ScienceQA}~\citeapp{SQA/lu2022learn} assesses multimodal science question answering. Models need to combine visual evidence, textual context, and scientific knowledge to answer multiple-choice science questions. We use \texttt{scienceqa\_img} for evaluations.

\item \textbf{VQAv2}~\citeapp{VQAv2/goyal2017making} evaluates general visual question answering on natural images. Its balanced question-answer design reduces language priors and tests robust image-question understanding. We use \texttt{vqav2\_test} for evaluations.

\item \textbf{TextVQA}~\citeapp{TextVQA/singh2019towards} focuses on text-rich visual question answering. It requires models to read scene text in images and reason over the recognized text to answer questions. We use \texttt{textvqa\_val} for evaluations and ocr's candidates are activated.

\item \textbf{OCRBench}~\citeapp{OCR-B/liu2024ocrbench} provides a targeted evaluation of OCR-related capabilities in large multimodal models. It covers text recognition, scene-text VQA, document-oriented VQA, key information extraction, and handwritten mathematical expression recognition.

\item \textbf{Vizwiz}~\citeapp{Vizwiz/gurari2018vizwiz} evaluates VQA in a real-world assistive setting. The images are captured by blind users and often contain blur, poor framing, ambiguity, or unanswerable questions, making the benchmark challenging for practical visual understanding. We use \texttt{vizwiz\_vqa\_val} for evaluations.

\item \textbf{SeedBench-Image}~\citeapp{SEED/Li_2024_CVPR} evaluates image-based multimodal comprehension with objective multiple-choice questions. It covers diverse capability dimensions and avoids subjective judge-based scoring.

\item\textbf{MMMU}~\citeapp{MMMU/yue2024mmmu} evaluates expert-level multimodal understanding and reasoning across multiple disciplines. It requires models to integrate visual perception with domain-specific knowledge and deliberate reasoning. We use \texttt{mmmu\_val} for evaluations.
\end{itemize}

\subsubsection{Video Benchmarks.}
We further evaluate LLaVA-Video-7B on four representative video-understanding benchmarks that cover temporal reasoning, long-form video understanding, egocentric perception, and objective video-based multimodal evaluation.
\begin{itemize}
\renewcommand\labelitemi{$\bullet$}
\item \textbf{MVBench}~\citeapp{MVBench/li2024mvbench} evaluates temporal understanding across diverse video tasks. Many tasks require dynamic information beyond a single frame, such as action recognition, temporal ordering, and event reasoning.

\item \textbf{Video-MME}~\citeapp{Video-MME/fu2025video} is a comprehensive video-understanding benchmark covering diverse video domains, durations, and multimodal cues. It evaluates both short- and long-context video comprehension and is suitable for assessing general video LVLMs.

\item \textbf{EgoSchema}~\citeapp{Ego-Schema/NEURIPS2023_90ce332a} focuses on long-form egocentric video question answering. Models answer multiple-choice questions based on three-minute first-person video clips, requiring long-range temporal reasoning over daily human activities. We use \texttt{egoschema\_subset} for comparison.

\item\textbf{SeedBench-Video}~\citeapp{SEED/Li_2024_CVPR} evaluates video-based multimodal understanding through objective multiple-choice questions. It complements MVBench and Video-MME by providing a structured evaluation of video comprehension abilities.
\end{itemize}

These benchmarks jointly test whether visual token pruning preserves performance across general image understanding, compositional reasoning, OCR, hallucination robustness, scientific and expert-level reasoning, real-world VQA, and temporal video comprehension.

\subsection{Implementation Details}
For performance evaluation, all experiments are conducted on four NVIDIA A6000 GPUs with a batch size of 1, while efficiency is measured on a single A6000 GPU. The implementation is based on Python 3.10, PyTorch 2.1.2, and CUDA 12.1. Following the empirical analysis in Sec.~\ref{app:empirical_analysis}, Cen-Prune applies the geometry correction after the multimodal projector in the LLaVA-series models and before the visual merger in Qwen2.5-VL. Before activating the diversity-based pruning mechanism, we compute the centered similarity matrix and the distinctiveness score, which introduces negligible computational overhead.

For the pruning ratios, we use 77.8\%, 88.9\%, and 94.4\% for both LLaVA-1.5 and LLaVA-NeXT, and 62.1\%, 81.1\%, and 90.5\% for LLaVA-Video. For LLaVA-NeXT, which processes high-resolution images using up to five crops ($5 \times 576 = 2880$ visual tokens), we follow the implementation of VisionZIP~\citeapp{VisionZip/YangCTWL0J25} and retain the same token proportion from each crop. For example, when an image is divided into three crops, the total number of visual tokens is $3 \times 576 = 1728$. To achieve a pruning ratio of 94.4\%, we retain 32 tokens from each crop, resulting in 96 retained tokens in total. For Qwen2.5-VL-7B, which supports dynamic-resolution inputs, we follow MMTok~\citeapp{MMTok/dong2026mmtok} and evaluate pruning ratios of 80\%, 90\%, and 95\%.

\subsection{Reproduction of Baselines and Integration}

\subsubsection{Baseline Reproducibility.}
We reproduce all baseline methods in a unified environment based on their original papers and official code repositories. For methods that do not report the 94.4\% pruning-ratio setting on LLaVA-1.5, we make minimal adaptations following their original configuration rules. We summarize the reproduction and integration details below.

\begin{itemize}
\renewcommand\labelitemi{$\bullet$}
\item \textbf{VisionZIP}~\citeapp{VisionZip/YangCTWL0J25}. For the setting with 32 retained visual tokens on LLaVA-1.5, we proportionally adapt its token allocation and retain 27 dominant tokens and 5 contextual tokens.

\item \textbf{HoloV}~\citeapp{HoloV/abs-2510-02912}. For the setting with 32 retained visual tokens on LLaVA-1.5, we follow its implementation and set $num_{crop}=[1024/N]$, where $N$ denotes the number of retained visual tokens.

\item \textbf{PruneSID}~\citeapp{PruneSID/fang2026prune}. For the setting with 32 retained visual tokens on LLaVA-1.5, we follow its default configuration and set the number of token groups to $K=\frac{N}{4}$, where $N$ denotes the number of retained visual tokens.

\item \textbf{VisPruner}~\citeapp{VisPruner/ZhangCLZZCGSZ25}. Following its default configuration, we set the important-token ratio to $r=0.5$ for all experiments.

\item \textbf{DivPrune}~\citeapp{DivPrune/AlvarSAZ25}. Following its default configuration, we apply token selection before the LLM. For fair comparison, visual token features are processed in \texttt{Float32} during pruning.

\item \textbf{ZOO-Prune}~\citeapp{ZOO-Prune/Kim_2026_CVPR}. Following its default configuration, we fix the perturbation hyperparameters to $m=64$ and $h=0.01$ for all experiments. For fair comparison, visual token features are processed in \texttt{Float32} during pruning.

\item \textbf{CDPruner}~\citeapp{CDPruner/zhang2026beyond}. Following its default configuration, we apply token selection before the LLM.

\item \textbf{MMTok}~\citeapp{MMTok/dong2026mmtok}. Following its default configuration, we set $\tau_t=0.02$, $\tau_v=0.2$, and $\alpha=0.5$ for all experiments on the LLaVA-series models. For Qwen2.5-VL-7B, we follow its implementation and reduce $\tau_t$ to $0.01$.
\end{itemize}

\input{Appendix/llava-NeXT-7b}

\input{Appendix/llava-NeXT-7b_HoloV}
\paragraph{Integration with Diversity-Based Pruners.}
In Sec.~\ref{sec:method}, we describe how Cen-Prune can be integrated into diversity-based pruning methods. We provide the implementation details below.

\begin{itemize}
\renewcommand\labelitemi{$\bullet$}
\item \textbf{DivPrune + Ours}. We replace the original selection geometry with the centered cosine similarity matrix. After DivPrune's original initialization, we multiply the minimum-distance score by our distinctiveness score at each subsequent step, and then select the next token by computing $\operatorname{argmax}(\text{dist\_scores} \times \text{distinctiveness})$.

\item \textbf{ZOO-Prune + Ours}. Similar to the integration with DivPrune, we replace the original selection geometry with the centered cosine similarity matrix and multiply ZOO-Prune's sensitivity score by our distinctiveness score. The two scores are normalized separately before multiplication.

\item \textbf{CDPruner + Ours}. Similar to the integration with ZOO-Prune, we multiply the original instruction-relevance score by our distinctiveness score, with each score normalized separately. The underlying kernel is replaced with the centered cosine similarity matrix.
\end{itemize}

\section{Further Quantitative Analysis and Results} \label{app:additional_results}

\subsection{Additional Quantitative Results}
\paragraph{Complete Results on LLaVA-NeXT-7B.} In the main paper, we reported the compression results of LLaVA-NeXT-7B when retaining up to 320 visual tokens. Tab.~\ref{tab:app_llava_next_7b_benchmarks} shows the complete results under all budget settings on LLaVA-NeXT-7B. Across different compression ratios, Cen-Prune's overall relative performance retention improved by 0.7\% to 2.0\% comparing to those diversity-based pruners.

\input{Appendix/llava-NeXT-13b}
\input{Appendix/llava-1.5-13b}

\input{Appendix/further_ablation}
\input{Appendix/Efficiency_llava-1.5-7b}

\paragraph{Different Protocol for Comparing with HoloV.} To ensure fairness in the comparison, on the high-resolution input architecture LLaVA-NeXT, we follow previous studies~\citeapp{VisionZip/YangCTWL0J25,VisPruner/ZhangCLZZCGSZ25,CDPruner/zhang2026beyond,PruneSID/fang2026prune,MMTok/dong2026mmtok} and perform pruning before removing padding in \texttt{AnyRes} branch. This means that each crop contains 576 tokens, up to a maximum of 2880 tokens, and is compressed at a fixed ratio per crop based on the pruning rate (77.8\% / 88.9\% / 94.4\%). For HoloV~\citeapp{HoloV/abs-2510-02912}, they apply visual token reduction after removing padding regions. We provide additional results in Tab.\ref{tab:app_llava_next_7b_holov} for comparing our Cen-Prune based on DivPrune~\citeapp{DivPrune/AlvarSAZ25} against HoloV with fixed 320 retained tokens. Under this setting, Cen-Prune still maintains a robust overall performance gain based on DivPrune, and also surpasses HoloV.

\paragraph{Additional Results on Larger-Scale Models.}
In the main paper, we report comparison results on 7B-scale models. To further validate the effectiveness of Cen-Prune across different model scales, we additionally compare it with existing baselines on LLaVA-NeXT-13B and LLaVA-1.5-13B. As shown in Tab.~\ref{tab:app_llava_next_13b_benchmarks} and Tab.~\ref{tab:app_llava_13b_benchmarks}, Cen-Prune can still deliver a robust overall performance improvement against diversity-based pruners.

\subsection{Additional Ablation Studies}
\paragraph{Ablation Studies on the Centering Strategy.}
We examine the effectiveness of mean-direction removal as an alternative centering strategy. Let $\bm{X}$ denote the working visual features. We remove the mean direction as follows:
$$
\bm{X}^{\perp}
=
\bm{X}
\left(
\bm{I}
-
\hat{\bm{\mu}}\hat{\bm{\mu}}^{\top}
\right),
\qquad
\hat{\bm{\mu}}
=
\frac{\bm{\mu}}{\lVert\bm{\mu}\rVert_2},
$$
where $\bm{\mu}$ is the mean feature vector computed over visual tokens. The first group in Tab.~\ref{tab:app_further_ablation} shows that completely projecting out the mean direction can discard informative token-specific variation along that direction. In contrast, mean centering removes only the shared offset while preserving relative variation among visual tokens.

\paragraph{Ablation Studies on the Distinctiveness Score.}
In the main paper, we compute the distinctiveness score from the raw cosine similarity matrix. We further conduct ablation studies with several score variants. As shown in the second group of Tab.~\ref{tab:app_further_ablation}, we first replace our cosine-based distinctiveness score with a Raw-$\ell_2$ score, defined as the mean unnormalized Euclidean distance between each token and all other tokens in the raw feature space. This modulus-aware distinctiveness metric improves performance on some benchmarks, especially the hallucination benchmark POPE. However, it leads to a noticeable performance drop on the text-oriented benchmark TextVQA. We also evaluate \texttt{[CLS]} attention with the same min--max normalization, but find that it is less compatible with our residual diversity objective.

\paragraph{Ablation Studies on Score Normalization.}
We further study the effect of score normalization in Tab.~\ref{tab:app_further_ablation}. Z-score normalization with a sigmoid transformation achieves performance comparable to our min--max normalization, with only minor differences across benchmarks. This suggests that the proposed distinctiveness score is robust to the normalization choice. We use min--max normalization in the main experiments for its simplicity and its slightly stronger average performance.

\subsection{Further Efficiency Results}
\paragraph{Complete Efficiency Comparison on LLaVA-1.5-7B.}
In the main paper, we provide the efficiency analysis on LLaVA-1.5-7B under the setting with 64 retained visual tokens. Here, we present the complete comparison under different pruning-ratio settings in Tab.~\ref{tab:app_efficiency_complete}. Across all token budgets, integrating Cen-Prune introduces negligible additional computational overhead. The small runtime differences observed across settings are within normal measurement variation and do not indicate a consistent change in inference cost.

\clearpage

\onecolumn

\section{Visualizations} \label{app:visualization}
\begin{figure*}[h]
  \centering
  \includegraphics[width=\textwidth]{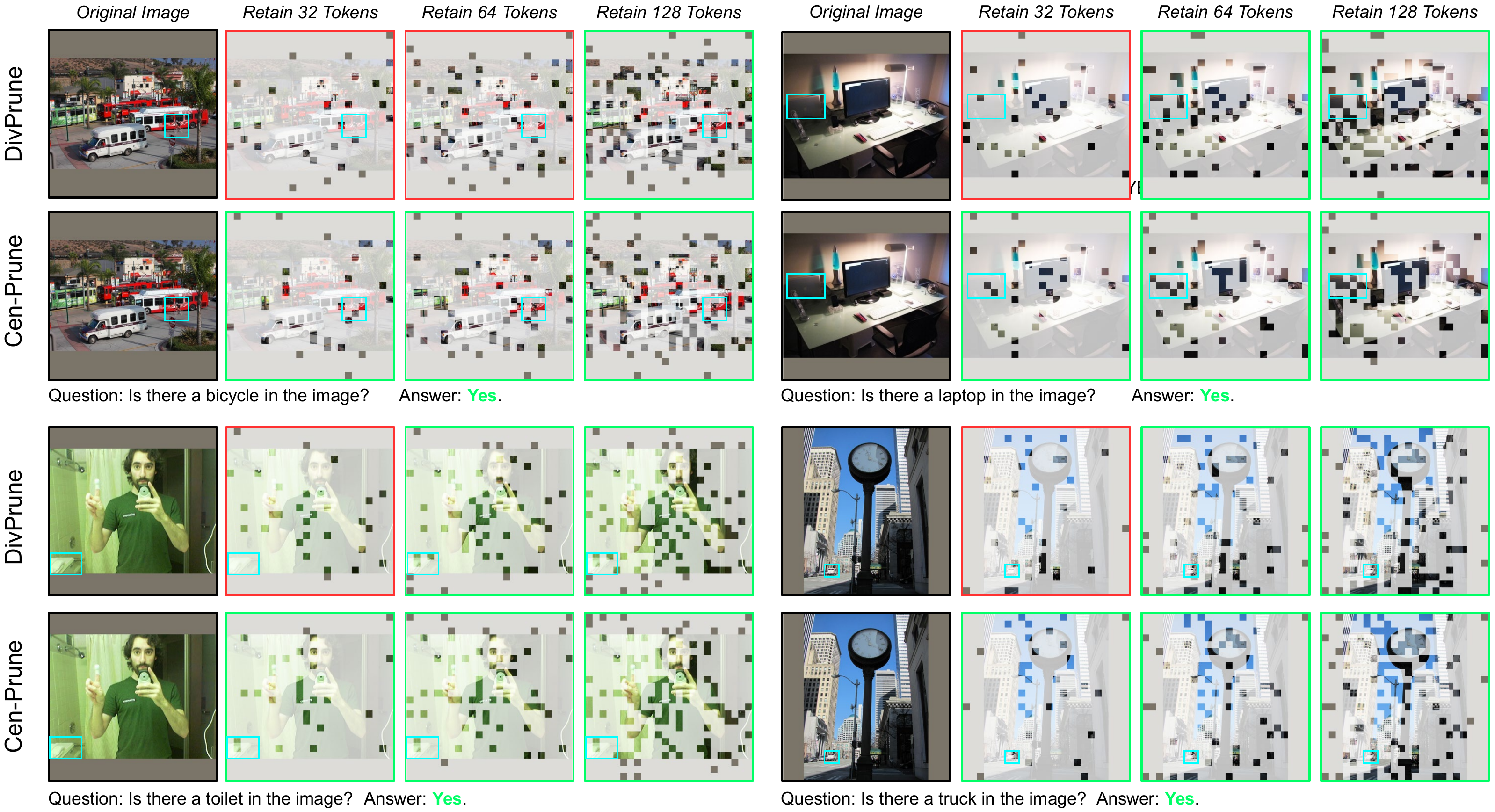}
  \caption{The case comparison of small object coverage between DivPrune and Cen-Prune from POPE.}
  \label{fig:app_visualization_pope}
  
  \centering
  \vspace{5pt}
  \includegraphics[width=\textwidth]{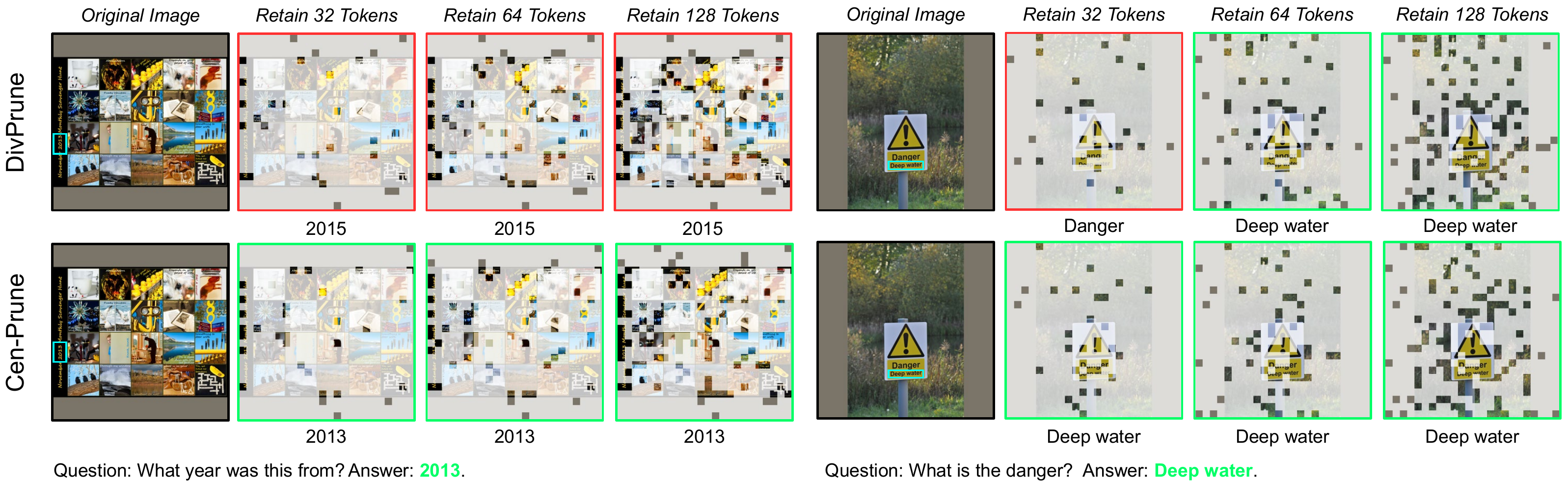}
  \caption{The case comparison of text-oriented Q\&A between DivPrune and Cen-Prune from TextVQA.}
  \label{fig:app_visualization_text}
  \vspace{-2pt}
\end{figure*}

We finally visualize the retained tokens under different reduction rates. As shown in Fig.~\ref{fig:app_visualization_pope}, a green outline indicates a correct answer, whereas a red outline indicates an incorrect answer. Compared with DivPrune, Cen-Prune retains more task-relevant visual cues, which is particularly noticeable under low-budget settings. When only 32 visual tokens are retained, Cen-Prune is often better able to cover small targets in the image, such as bicycles and trucks.

As shown in Fig.~\ref{fig:app_visualization_text}, we further compare Cen-Prune with DivPrune on text-oriented images. Even with a relatively relaxed budget of 128 retained visual tokens, DivPrune struggles to preserve the integrity of textual content, which can lead to incomplete text recognition.

\twocolumn


%% file: appendix_A.tex
\section{Details of Empirical Analysis}
\label{app:empirical_analysis}

\subsection{Data and Protocol for Geometry Analysis}

All our analyses in Sec. \ref{sec:analysis} are based on the same fixed set of 1{,}000 \textit{probe} images sampled from COCO val2017~\cite{COCO/lin2014microsoft} with seed 0. The resulting probe list is frozen and shared across all backbones and feature spaces. For each image, we extract the complete visual-token matrix
$\bm{X}=[\bm{x}_1,\ldots,\bm{x}_N]^\top\in\mathbb{R}^{N\times D}$
at the feature space under study. Unless otherwise specified, each statistic is computed per image and summarized by its unweighted mean and standard deviation across the probe set. Pairwise statistics exclude self-pairs and count each unordered pair once.

For LLaVA-1.5-7B, we examine the exact input and output of the multimodal projector. For Qwen2.5-VL-7B-Instruct, we examine the encoder representation consumed by the pruning implementation and the post-merger representation passed to the language model. Following the feature construction used by MMTok~\cite{MMTok/dong2026mmtok}, the former averages each merger-aligned group of four encoder patch features and restores the resulting tokens to raster order. Because Qwen processes images at dynamic resolution, its token count varies across images.

\subsection{Quantities and Metrics}
\label{app:diagnostics}

Let
$\hat{\bm{x}}_i=\bm{x}_i/\lVert\bm{x}_i\rVert_2$
and
$\bm{S}=\hat{\bm{X}}\hat{\bm{X}}^\top$.

\paragraph{Mean off-diagonal cosine.}
We define
\[
\bar{s}
=
\frac{1}{N(N-1)}
\sum_{i\neq j}S_{ij},
\]
which measures the average directional agreement between distinct token pairs. This metric has been used in Tab.~\ref{tab:selector_performance}.

\paragraph{Negative-pair fraction.}
We denote by $\rho_-$ the fraction of distinct token pairs satisfying $S_{ij}<0$.  This metric has been used in Tab.~\ref{tab:selector_performance}.

\paragraph{Effective rank.}
Let $\lambda_k$ be the eigenvalues of the cosine Gram matrix $\bm S$ and
$p_k=\lambda_k/\sum_\ell\lambda_\ell$. We define
\[
\operatorname{erank}(\bm S)
=
\exp\left(-\sum_k p_k\log p_k\right),
\]
which has been studied in previous works~\cite{erank/roy2007effective,erank/giraldo2014measures}.
Because $\bm S$ is constructed from row-normalized features, this quantity measures directional spread independently of token magnitude. It equals one when all normalized tokens lie in a single one-dimensional subspace. Unless stated otherwise, we report the effective rank of the uncentered cosine Gram matrix, which characterizes the geometry available before applying Cen-Prune. We do not assume that effective rank must increase monotonically under centering.

\subsection{Where the Correction Applies}

Cen-Prune is defined for a generic visual-token representation, but the appropriate feature space depends on the backbone. Table~\ref{tab:space_geometry} compares the geometries immediately before and after the corresponding cross-modal projection stage.

In LLaVA-1.5, the multimodal projector increases concentration: the mean cosine rises from $0.309$ to $0.366$, the negative-pair fraction falls from $3.57\%$ to $2.27\%$, and effective rank decreases from $21.79$ to $13.20$. Thus, although the pre-projector representation is already positively biased, the post-projector geometry provides less directional contrast.

Qwen2.5-VL exhibits the opposite progression. Its merger-aligned encoder representation is nearly collinear, with a mean cosine of $0.993$, no observed negative pairs, and an effective rank of $1.06$. The visual merger substantially disperses this geometry: the mean cosine decreases to $0.300$, while effective rank increases to $31.31$.

We therefore apply Cen-Prune to the post-projector representation in LLaVA-1.5 and to the merger-aligned encoder representation in Qwen2.5-VL. These are the most concentrated feature spaces consumed by the corresponding pruning pipelines.

\begin{table*}[t]
  \centering
  \small
  \begin{tabular}{llccccc}
  \toprule
  Backbone & Space & $N$ & $D$ & $\bar{s}$ &
  $\rho_-$ (\%) & $\operatorname{erank}(\bm{S})$ \\
  \midrule
  LLaVA-1.5
  & Pre-projector
  & 576 & 1024
  & $0.309{\pm}0.040$
  & $3.57{\pm}2.31$
  & $21.79{\pm}5.96$ \\

  LLaVA-1.5
  & Post-projector
  & 576 & 4096
  & $0.366{\pm}0.025$
  & $2.27{\pm}0.86$
  & $13.20{\pm}2.38$ \\

  Qwen2.5-VL
  & Pre-merger (encoder average)
  & 345 [88, 529] & 1280
  & $0.993{\pm}0.003$
  & $0.00{\pm}0.00$
  & $1.060{\pm}0.016$ \\

  Qwen2.5-VL
  & Post-merger
  & 345 [88, 529] & 3584
  & $0.300{\pm}0.030$
  & $0.48{\pm}0.29$
  & $31.31{\pm}5.57$ \\
  \bottomrule
  \end{tabular}
  \caption{Geometry of the feature spaces examined in LLaVA-1.5-7B and Qwen2.5-VL-7B-Instruct. Values are image-level mean $\pm$ standard deviation over the fixed probe set. For Qwen, $N$ is reported as median [minimum, maximum]. Effective rank is computed from the uncentered cosine Gram matrix.}
  \label{tab:space_geometry}
  \vspace{-2pt}
\end{table*}

\subsection{Coordinate Conditioning and Score Normalization}
\label{app:qwen_geometry}

Centering is the principal geometric correction in Cen-Prune. When variance is reasonably balanced across feature dimensions, as in the LLaVA projector representation, centering can be applied to the raw visual-token matrix $\bm V=[\bm v_1,\ldots,\bm v_N]^\top$. In some backbones, however, a small number of dimensions account for most of the feature variance and consequently dominate the cosine geometry. Although centering still removes the shared component in such cases, it does not prevent the residual similarities from being governed by these high-variance dimensions.

\begin{figure*}[t]
  \centering
  \includegraphics[width=\textwidth]{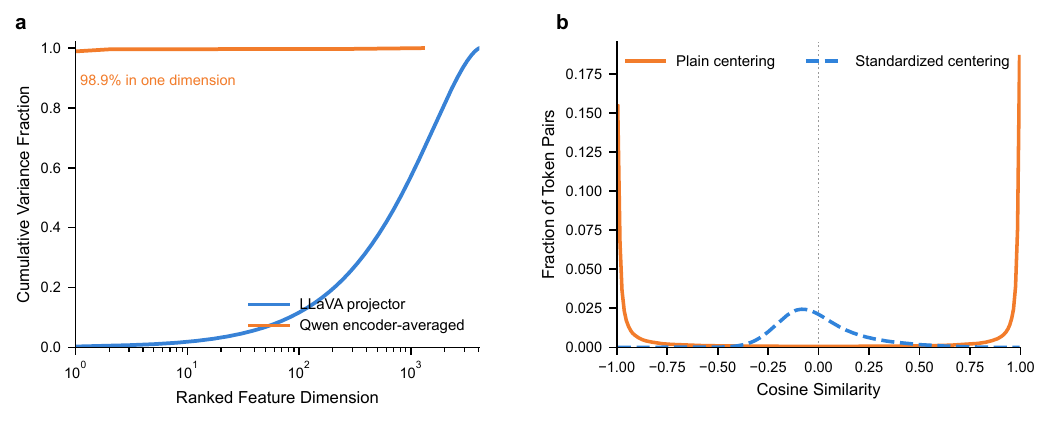}
  \caption{\textbf{Standardized centering corrects Qwen's variance-dominated residual geometry.} \textbf{(a)} Mean cumulative fraction of centered feature variance, with dimensions ranked by variance independently for each image. The highest-variance dimension accounts for $98.9\%$ of the total centered variance in Qwen's merger-aligned encoder representation, compared with only $0.28\%$ in the LLaVA projector representation. Shading denotes one standard deviation across images. \textbf{(b)} Pair-pooled distributions of off-diagonal cosine similarities for Qwen after plain and standardized centering. Standardized centering largely eliminates the endpoint concentration that remains after plain centering. The curves are normalized independently and linearly interpolated from bins of width $0.01$ for visualization.}
  \label{fig:qwen_standardization}
  \vspace{-2pt}
\end{figure*}

Qwen's merger-aligned encoder representation is an example of this variance imbalnace. After plain centering, $35.7\%$ of token pairs have cosine similarity below $-0.9$, while $38.9\%$ have similarity above $0.9$; thus, $74.6\%$ satisfy $|\cos(\bm{x}_i,\bm{x}_j)|>0.9$. This concentration near the two endpoints indicates that centering along does not yield a well-conditioned residual geometry. As shown in Fig.~\ref{fig:qwen_standardization}a, the highest-variance dimension accounts for $98.9\%$ of the centered variance on average, and the ten highest-variance dimensions collectively account for $99.6\%$. By comparison, the corresponding proportions in the LLaVA projector representation are only $0.28\%$ and $1.86\%$, respectively.
Consequently, inner products between Qwen's centered tokens are determined largely by the sign of a single coordinate, and subsequent $\ell_2$-normalization pushes their cosine similarities toward either $-1$ or $+1$.

We therefore apply per-dimension variance scaling before performing the same image-wise centering operation. For each image, let
\begin{equation}
\bm Y
=
\bm V\operatorname{diag}(\bm\sigma)^{-1},
\quad
\sigma_d
=
\left(
\frac{1}{N}\sum_{i=1}^{N}(v_{id}-\mu_d)^2+\epsilon
\right)^{1/2},
\end{equation}
where $\bm\mu=N^{-1}\sum_i\bm v_i$. Both $\bm\mu$ and $\bm\sigma$ are computed from the tokens of the current image; no statistics are shared across images or estimated from external data. Cen-Prune then centers the variance-scaled working coordinates $\bm Y$:
\begin{equation}
\bar{\bm Y}
=
\bm Y-\bm 1\bm\mu_Y^\top
=
(\bm V-\bm 1\bm\mu^\top)
\operatorname{diag}(\bm\sigma)^{-1},
\quad
\bm\mu_Y
=
\frac{1}{N}\sum_i\bm y_i.
\end{equation}
Thus, per-dimension scaling does not replace mean centering; it defines the coordinate system in which centering is performed. The uncentered working coordinates $\bm Y$ are retained for the distinctiveness branch, whereas only their mean-removed form $\bar{\bm Y}$ is used to construct the pairwise diversity geometry. If $\bm\sigma$ is constant across dimensions, this construction differs from plain centering only by a global scale and therefore produces the same cosine-similarity matrix.

This standardized centering reduces the fraction of Qwen token pairs satisfying $\lvert\cos(\bm x_i,\bm x_j)\rvert>0.9$ from $74.6\%$ to $0.008\%$ and increases the normalized-spectrum effective rank from $1.73$ to $53.65$ (Fig.~\ref{fig:qwen_standardization}b). By comparison, variance is substantially more balanced across the feature dimensions of LLaVA: standardization changes the corresponding fraction only from $7.38\%$ to $6.69\%$, although it still moderately increases the effective rank. These measurements motivate plain centering for LLaVA and standardized centering for Qwen. Both variants preserve the principal operation of Cen-Prune: removing the image-wise mean before constructing the pairwise diversity matrix.

The raw distinctiveness scores must also be normalized before they can serve as nonnegative multiplicative weights. Given the scores $\{g_i\}_{i=1}^{N}$, we consider two mappings applied independently to each image. Min--max normalization gives
\begin{equation}
\tilde g_i^{\mathrm{mm}}
=
\frac{g_i-g_{\min}}
     {g_{\max}-g_{\min}+\epsilon},
\end{equation}
which preserves the score ordering and maps the scores to the unit interval. Its relative scale is determined by the two extrema within each image. We use this mapping for LLaVA, whose token count is fixed at $N=576$. For Qwen, dynamic image resolution changes the number and composition of the visual tokens and, consequently, the extrema observed in each image. We therefore apply z-score normalization followed by a sigmoid:
\begin{equation}
\tilde g_i^{\mathrm{zn}}
=
\operatorname{sigmoid}
\left(
\frac{g_i-\mu_g}{\sigma_g+\epsilon}
\right),
\end{equation}
where $\mu_g$ and $\sigma_g$ are computed over the tokens of the same image. This mapping remains monotonic, compresses extreme values, and expresses each score relative to the image-wise score distribution rather than directly to its extrema. Thus, neither mapping changes the within-image token ordering, but each controls the relative strength of the multiplicative distinctiveness weighting differently. We use min--max normalization for LLaVA and sigmoid-transformed z scores for Qwen. Each choice is fixed for its respective backbone and shared across all datasets and retention budgets, with no normalization statistics transferred across images.

%% file: Appendix/llava-NeXT-7b.tex
\begin{table*}[t]
\centering

\setlength{\tabcolsep}{3.5pt} 
\renewcommand{\arraystretch}{1.0} 

\footnotesize
\scalebox{0.8}{
\begin{tabular}{l cccccccccc c}
\toprule[1.2pt]
\textbf{Method} & \textbf{GQA} & \textbf{MMB} & \textbf{MME} & \textbf{POPE} & \textbf{SQA} & \textbf{VQA\textsuperscript{Text}} & \textbf{OCR-B} & \textbf{VizWiz} & \textbf{SEED-I} & \textbf{MMMU} & \textbf{Avg.} \\
\midrule
\rowcolor{gray!10} \textit{Upper Bound (2880 Tokens)} & 64.2 & 67.9 & 1520 & 86.4 & 70.2 & 61.3 & 51.3 & 59.7 & 69.6 & 37.2 & 100.0\% \\
\midrule

\rowcolor{gray!10} \textbf{LLaVA-NeXT-7B} & \multicolumn{10}{c}{\textit{Retain Up to 640 Tokens} \textbf{\textcolor{mydarkgreen}{($\downarrow$ 77.8\%)}}} & \\ 
VisionZip\confyear{CVPR25} & 61.2 & 65.0 & 1474 & 86.1 & 68.0 & 60.3 & 47.8 & 61.0 & 66.7 & 36.6 & 97.2\% \\
VisPruner\confyear{ICCV25} & 61.6 & 64.9 & 1472 & 86.0 & 67.8 & 59.8 & 45.1 & 60.4 & 66.4 & 36.1 & 96.3\% \\
PruneSID\confyear{ICLR26} & 61.6 & 63.9 & 1473 & 86.2 & 68.1 & 55.5 & 34.3 & 59.2 & 67.0 & 36.4 & 93.4\% \\

DivPrune\confyear{CVPR25} & 62.0 & 65.4 & 1496 & 87.1 & 67.8 & 57.0 & 40.0 & 59.3 & 67.6 & 36.1 & 95.3\% \\
\rowcolor{mysoftgreen} \quad \textbf{+ Ours} & 62.3 & 66.0 & 1528 & 87.1 & 68.1 & 58.2 & 44.1 & 59.6 & 68.2 & 36.3 & 96.9\% \\
ZOO-Prune\confyear{CVPR26} & 61.8 & 64.9 & 1488 & 86.8 & 68.1 & 58.1 & 43.8 & 59.3 & 67.6 & 36.9 & 96.3\% \\
\rowcolor{mysoftgreen} \quad \textbf{+ Ours} & 62.3 & 65.8 & 1495 & 86.9 & 67.8 & 59.2 & 47.0 & 59.5 & 68.2 & 35.1 & 97.0\% \\
\categoryrule
MMTok\confyear{ICLR26} & 62.0 & 65.4 & 1507 & 87.0 & 68.2 & 59.1 & 44.8 & 59.2 & 67.6 & 36.8 & 96.9\% \\
CDPruner\confyear{NeurIPS25} & 62.7 & 65.9 & 1495 & 87.4 & 68.0 & 58.8 & 44.7 & 59.0 & 69.0 & 36.3 & 97.0\% \\
\rowcolor{mysoftgreen} \quad \textbf{+ Ours} & 62.4 & 66.1 & 1508 & 87.2 & 67.5 & 60.3 & 48.4 & 59.4 & 68.9 & 36.9 & \textbf{98.1\%} \\
\midrule
\rowcolor{gray!10} \textbf{LLaVA-NeXT-7B} & \multicolumn{10}{c}{\textit{Retain Up to 320 Tokens} \textbf{\textcolor{mydarkgreen}{($\downarrow$ 88.9\%)}}} & \\ 
VisionZip\confyear{CVPR25} & 58.9 & 63.4 & 1401 & 82.1 & 67.3 & 58.8 & 39.2 & 59.9 & 63.5 & 36.7 & 93.0\% \\
VisPruner\confyear{ICCV25} & 58.6 & 63.7 & 1404 & 81.1 & 67.9 & 58.6 & 37.5 & 61.1 & 62.8 & 35.7 & 92.5\% \\
PruneSID\confyear{ICLR26} & 60.6 & 63.0 & 1468 & 85.0 & 67.8 & 54.3 & 32.4 & 58.7 & 65.8 & 36.9 & 92.2\% \\

DivPrune\confyear{CVPR25} & 61.1 & 63.7 & 1375 & 84.6 & 67.8 & 56.3 & 34.2 & 59.7 & 65.7 & 36.7 & 92.5\% \\
\rowcolor{mysoftgreen} \quad \textbf{+ Ours} & 60.9 & 64.6 & 1480 & 85.1 & 68.0 & 57.6 & 39.3 & 59.2 & 66.4 & 36.2 & 94.5\% \\
ZOO-Prune\confyear{CVPR26} & 61.0 & 64.1 & 1413 & 85.3 & 67.4 & 57.1 & 36.7 & 58.9 & 65.8 & 37.2 & 93.5\% \\
\rowcolor{mysoftgreen} \quad \textbf{+ Ours} & 60.8 & 64.6 & 1443 & 85.4 & 67.8 & 58.0 & 39.6 & 58.9 & 66.2 & 36.3 & 94.3\% \\
\categoryrule
MMTok\confyear{ICLR26} & 61.1 & 64.1 & 1471 & 85.8 & 67.6 & 56.9 & 34.6 & 59.1 & 66.2 & 38.0 & 93.8\% \\
CDPruner\confyear{NeurIPS25} & 61.6 & 64.2 & 1485 & 87.3 & 66.8 & 57.0 & 38.8 & 58.4 & 67.3 & 35.8 & 94.4\% \\
\rowcolor{mysoftgreen} \quad \textbf{+ Ours} & 61.4 & 65.3 & 1490 & 86.8 & 67.2 & 58.9 & 43.3 & 58.8 & 67.7 & 35.9 & \textbf{95.8\%} \\
\midrule

\rowcolor{gray!10} \textbf{LLaVA-NeXT-7B} & \multicolumn{10}{c}{\textit{Retain Up to 160 Tokens} \textbf{\textcolor{mydarkgreen}{($\downarrow$ 94.4\%)}}} & \\ 
VisionZip\confyear{CVPR25} & 55.5 & 59.9 & 1303 & 74.9 & 68.3 & 56.3 & 30.5 & 60.4 & 58.4 & 36.3 & 87.8\% \\
VisPruner\confyear{ICCV25} & 56.5&59.7&1290&75.2&69.0&	56.0&30.7&60.1&58.2&35.4 &87.7\%\\
PruneSID\confyear{ICLR26} & 58.6 & 60.9 & 1389 & 80.1 & 67.3 & 52.5 & 27.4 & 58.8 & 62.2 & 36.7 & 88.6\% \\
DivPrune\confyear{CVPR25} & 59.2 & 63.1 & 1375 & 79.4 & 67.6 & 54.7 & 29.4 & 59.6 & 62.9 & 36.7 & 89.9\% \\
\rowcolor{mysoftgreen} \quad \textbf{+ Ours} & 60.0 & 63.7 & 1392 & 82.6 & 68.2 & 56.2 & 32.3 & 60.0 & 63.9 & 35.6 & 91.4\% \\
ZOO-Prune\confyear{CVPR26} & 59.4 & 62.4 & 1367 & 81.7 & 68.8 & 55.3 & 30.7 & 60.3 & 62.8 & 36.2 & 90.5\% \\
\rowcolor{mysoftgreen} \quad \textbf{+ Ours} & 59.5 & 63.6 & 1388 & 82.7 & 68.2 & 56.1 & 32.8 & 60.4 & 64.1 & 37.2 & 91.9\% \\
\categoryrule
MMTok\confyear{ICLR26} & 60.0 & 63.7 & 1413 & 83.8 & 68.0 & 54.5 & 30.1 & 59.3 & 64.9 & 37.1 & 91.4\% \\
CDPruner\confyear{NeurIPS25} & 60.9 & 64.7 & 1422 & 87.0 & 66.9 & 55.4 & 32.9 & 58.3 & 65.7 & 35.4 & 92.1\% \\
\rowcolor{mysoftgreen} \quad \textbf{+ Ours} & 60.8 & 65.1 & 1436 & 86.2 & 66.9 & 56.8 & 37.0 & 58.5 & 66.3 & 35.8 & \textbf{93.4\%} \\

\bottomrule[1.2pt]
\end{tabular}%
}
\vspace{-5pt}
\caption{Evaluation on image benchmarks for LLaVA-NeXT-7B under different token reduction ratio.}
\vspace{-2pt}
\label{tab:app_llava_next_7b_benchmarks}
\end{table*}

%% file: Appendix/llava-NeXT-7b_HoloV.tex
\begin{table*}[t]
\centering

\setlength{\tabcolsep}{3.5pt} 
\renewcommand{\arraystretch}{1.0} 

\footnotesize
\scalebox{0.8}{
\begin{tabular}{l cccccccccc c}
\toprule[1.2pt]
\textbf{Method} & \textbf{GQA} & \textbf{MMB} & \textbf{MME} & \textbf{POPE} & \textbf{SQA} & \textbf{VQA\textsuperscript{Text}} & \textbf{OCR-B} & \textbf{VizWiz} & \textbf{SEED-I} & \textbf{MMMU} & \textbf{Avg.} \\
\midrule
\rowcolor{gray!10} \textit{Upper Bound (2880 Tokens)} & 64.2 & 67.9 & 1520 & 86.4 & 70.2 & 61.3 & 51.3 & 59.7 & 69.6 & 37.2 & 100.0\% \\
\midrule

\rowcolor{gray!10} \textbf{LLaVA-NeXT-7B} & \multicolumn{10}{c}{\textit{Retain 320 Tokens}} & \\ 
HoloV\confyear{NeurIPS25} & 59.4 & 65.5 & 1438 & 83.1 & 66.8 & 55.6 & 36.6 & 57.8 & 64.7 & 35.4 & 92.2\% \\
DivPrune\confyear{CVPR25} & 60.4 & 64.3 & 1439 & 84.8 & 68.0 & 54.0 & 35.9 & 57.6 & 66.1 & 37.8 & 93.0\% \\
\rowcolor{mysoftgreen} \quad \textbf{+ Ours} & 61.0 & 65.0 & 1477 & 85.7 & 67.8 & 56.9 & 40.3 & 58.1 & 66.9 & 36.9 & \textbf{94.8}\% \\
\bottomrule[1.2pt]
\end{tabular}%
}
\vspace{-5pt}
\caption{Performance comparison on LLaVA-NeXT-7B under a fixed budget setting.}
\vspace{-5pt}
\label{tab:app_llava_next_7b_holov}
\end{table*}

%% file: Appendix/llava-NeXT-13b.tex
\begin{table*}[t]
\centering

\setlength{\tabcolsep}{3.5pt} 
\renewcommand{\arraystretch}{1.0} 

\footnotesize
\scalebox{0.8}{
\begin{tabular}{l cccccccccc c}
\toprule[1.2pt]
\textbf{Method} & \textbf{GQA} & \textbf{MMB} & \textbf{MME} & \textbf{POPE} & \textbf{SQA} & \textbf{VQA\textsuperscript{Text}} & \textbf{OCR-B} & \textbf{VizWiz} & \textbf{SEED-I} & \textbf{MMMU} & \textbf{Avg.} \\
\midrule
\rowcolor{gray!10} \textit{Upper Bound (2880 Tokens)} & 64.4 & 67.7 & 1551 & 85.1 & 73.0 & 63.3 & 54.5 & 62.4 & 71.6 & 36.4 & 100.0\% \\
\midrule

\rowcolor{gray!10} \textbf{LLaVA-NeXT-7B} & \multicolumn{10}{c}{\textit{Retain Up to 640 Tokens} \textbf{\textcolor{mydarkgreen}{($\downarrow$ 77.8\%)}}} & \\ 
DivPrune\confyear{CVPR25} & 60.0  &  64.3  &  1484  &  81.6  &  71.9  &  56.1  &  30.4  &  58.1  &  64.4  &  36.2 & 90.5\% \\
\rowcolor{mysoftgreen} \quad \textbf{+ Ours} & 61.2  &  65.3  &  1472  &  83.5  &  72.1  &  58.3  &  34.7  &  59.4  &  65.7  &  36.3 & \textbf{92.6\%} \\

\midrule
\rowcolor{gray!10} \textbf{LLaVA-NeXT-7B} & \multicolumn{10}{c}{\textit{Retain Up to 320 Tokens} \textbf{\textcolor{mydarkgreen}{($\downarrow$ 88.9\%)}}} & \\ 
DivPrune\confyear{CVPR25} & 61.9  &  65.2  &  1498  &  85.1  &  72.7  &  57.9  &  35.8  &  60.0  &  67.2  &  37.2 & 93.8\% \\
\rowcolor{mysoftgreen} \quad \textbf{+ Ours} & 62.8 &  66.2 &  1522 &  85.3 &  73.0 &  59.8 &  40.2 &  60.3 &  68.2 &  36.7 &  \textbf{95.5\%} \\

\midrule
\rowcolor{gray!10} \textbf{LLaVA-NeXT-7B} & \multicolumn{10}{c}{\textit{Retain Up to 160 Tokens} \textbf{\textcolor{mydarkgreen}{($\downarrow$ 94.4\%)}}} & \\ 
DivPrune\confyear{CVPR25} & 63.7 &  68.0 &  1521 &  86.4 &  72.9 &  59.1 &  43.7 &  60.5 &  69.7 &  37.8 & 97.0\%
 \\
\rowcolor{mysoftgreen} \quad \textbf{+ Ours} & 64.0 &  68.0 &  1554 &  86.5 &  72.7 &  61.2 &  46.9 &  61.8 &  70.2 &  36.2 & \textbf{98.0\%} \\

\bottomrule[1.2pt]
\end{tabular}%
}
\vspace{-5pt}
\caption{Comparison on LLaVA-NeXT-13B under different token reduction ratio.}
\vspace{-2pt}
\label{tab:app_llava_next_13b_benchmarks}
\end{table*}

%% file: Appendix/llava-1.5-13b.tex
\renewcommand{\multirowsetup}{\centering}
\begin{table*}[!ht]
\centering

\setlength{\tabcolsep}{3.5pt} 
\renewcommand{\arraystretch}{1.0} 

\footnotesize
\scalebox{0.8}{
\begin{tabular}{l cccccccccc c}
\toprule[1.2pt]
\textbf{Method} & \textbf{GQA} & \textbf{MMB} & \textbf{MME} & \textbf{POPE} & \textbf{SQA} & \textbf{VQA\textsuperscript{Text}} & \textbf{OCR-B} & \textbf{VizWiz} & \textbf{SEED-I} & \textbf{MMMU} & \textbf{Avg.} \\
\midrule
\rowcolor{gray!10} \textit{Upper Bound (576 Tokens)} & 63.2 & 67.7 & 1522 & 85.9 & 72.8 & 61.3 & 33.6 & 56.6 & 66.9 & 36.4 & 100.0\% \\
\midrule

\rowcolor{gray!10} \textbf{LLaVA-1.5-13B} & \multicolumn{10}{c}{\textit{Retain 128 Tokens} \textbf{\textcolor{mydarkgreen}{($\downarrow$ 77.8\%)}}}  & \multicolumn{1}{c}{} \\ 
VisionZip\confyear{CVPR25} & 57.8 & 67.0 & 1458 & 82.4 & 74.0 & 58.7 & 32.8 & 55.1 & 63.7 & 35.8 & 96.8\% \\
PruneSID\confyear{ICLR26} & 58.4 & 65.4 & 1445 & 83.9 & 72.9 & 57.5 & 29.3 & 55.9 & 64.1 & 36.7 & 95.8\% \\

DivPrune\confyear{CVPR25} & 59.0 & 66.8 & 1446 & 86.7 & 72.8 & 58.0 & 30.5 & 56.5 & 63.9 & 36.4 & 96.9\% \\
\rowcolor{mysoftgreen} \quad \textbf{+ Ours} & 59.5 & 67.2 & 1470 & 86.4 & 73.2 & 59.0 & 32.9 & 56.1 & 64.5 & 35.9 & 97.9\% \\ 
ZOO-Prune\confyear{CVPR26} & 58.9 & 66.5 & 1480 & 86.8 & 73.5 & 58.7 & 32.2 & 55.7 & 64.8 & 35.3 & 97.4\% \\
\rowcolor{mysoftgreen} \quad \textbf{+ Ours} & 59.2 & 67.0 & 1460 & 86.3 & 73.4 & 59.2 & 32.5 & 55.8 & 64.8 & 35.9 & 97.7\% \\
\categoryrule
MMTok\confyear{ICLR26} & 59.1 & 66.8 & 1434 & 86.2 & 73.5 & 58.8 & 33.6 & 56.0 & 64.6 & 35.3 & 97.6\% \\
CDPruner\confyear{NeurIPS25} & 59.8 & 67.0 & 1448 & 87.2 & 72.7 & 58.6 & 32.1 & 55.8 & 65.0 & 36.0 & 97.6\% \\
\rowcolor{mysoftgreen} \quad \textbf{+ Ours} & 60.0 & 67.5 & 1447 & 86.3 & 72.9 & 58.9 & 33.7 & 55.5 & 65.1 & 36.3 & \textbf{98.2\%} \\

\midrule

\rowcolor{gray!10} \textbf{LLaVA-1.5-13B} & \multicolumn{10}{c}{\textit{Retain 64 Tokens} \textbf{\textcolor{mydarkgreen}{($\downarrow$ 88.9\%)}}} & \multicolumn{1}{c}{} \\
VisionZip\confyear{CVPR25} & 56.1 & 64.8 & 1403 & 75.9 & 74.2 & 57.5 & 31.0 & 56.1 & 60.2 & 36.1 & 94.1\% \\
PruneSID\confyear{ICLR26} & 57.9 & 63.9 & 1437 & 82.6 & 71.5 & 56.7 & 28.6 & 57.7 & 62.4 & 35.6 & 94.6\% \\

DivPrune\confyear{CVPR25} & 57.4 & 64.6 & 1476 & 84.7 & 71.8 & 57.1 & 30.1 & 58.1 & 62.3 & 35.0 & 95.5\% \\
\rowcolor{mysoftgreen} \quad \textbf{+ Ours} & 58.3 & 64.4 & 1466 & 84.6 & 72.3 & 58.2 & 31.3 & 57.8 & 63.3 & 35.8 & 96.5\% \\
ZOO-Prune\confyear{CVPR26} & 58.8 & 64.4 & 1460 & 85.7 & 71.9 & 58.1 & 29.9 & 57.5 & 63.1 & 36.2 & 96.2\% \\
\rowcolor{mysoftgreen} \quad \textbf{+ Ours} & 58.9 & 66.0 & 1478 & 85.2 & 72.8 & 58.5 & 32.2 & 57.5 & 63.4 & 36.3 & 97.4\% \\
\categoryrule
MMTok\confyear{ICLR26} & 58.5 & 65.7 & 1464 & 84.5 & 72.8 & 58.0 & 31.8 & 57.6 & 63.6 & 35.2 & 96.7\% \\
CDPruner\confyear{NeurIPS25} & 59.3 & 65.5 & 1442 & 87.3 & 72.1 & 57.6 & 30.1 & 56.6 & 64.0 & 34.4 & 96.0\% \\
\rowcolor{mysoftgreen} \quad \textbf{+ Ours} & 59.1 & 66.6 & 1472 & 86.4 & 71.8 & 58.7 & 32.2 & 57.3 & 64.4 & 36.0 & \textbf{97.6\%} \\
\midrule

\rowcolor{gray!10} \textbf{LLaVA-1.5-13B} & \multicolumn{10}{c}{\textit{Retain 32 Tokens} \textbf{\textcolor{mydarkgreen}{($\downarrow$ 94.4\%)}}} & \multicolumn{1}{c}{} \\ 
VisionZip\confyear{CVPR25} & 58.3 & 61.5 & 1270 & 67.0 & 72.3 & 55.3 & 27.9 & 57.1 & 55.9 & 35.1 & 89.8\% \\
PruneSID\confyear{ICLR26} & 55.6 & 61.9 & 1335 & 77.2 & 71.8 & 54.5 & 26.1 & 58.2 & 59.4 & 35.7 & 91.2\% \\

DivPrune\confyear{CVPR25} & 56.0 & 62.1 & 1380 & 79.7 & 70.8 & 54.6 & 26.4 & 57.8 & 59.6 & 34.6 & 91.5\% \\
\rowcolor{mysoftgreen} \quad \textbf{+ Ours} & 57.0 & 63.2 & 1446 & 80.8 & 72.4 & 56.3 & 29.0 & 58.4 & 61.0 & 36.1 & 94.4\% \\

ZOO-Prune\confyear{CVPR26} & 57.0 & 63.6 & 1423 & 81.6 & 72.3 & 56.4 & 28.8 & 58.4 & 60.7 & 35.8 & 94.2\% \\
\rowcolor{mysoftgreen} \quad \textbf{+ Ours} & 57.4 & 64.3 & 1422 & 81.6 & 72.3 & 57.3 & 29.0 & 58.0 & 61.3 & 35.9 & 94.6\% \\
\categoryrule
MMTok\confyear{ICLR26} & 57.8 & 63.7 & 1427 & 82.4 & 73.2 & 55.7 & 28.2 & 58.5 & 61.7 & 35.0 & 94.2\% \\
CDPruner\confyear{NeurIPS25} & 58.5 & 64.0 & 1418 & 87.5 & 72.3 & 55.5 & 27.0 & 56.9 & 62.8 & 35.8 & 94.5\% \\
\rowcolor{mysoftgreen} \quad \textbf{+ Ours} & 58.6 & 65.6 & 1427 & 85.5 & 72.6 & 57.4 & 29.6 & 57.1 & 63.2 & 35.1 & \textbf{95.6\%} \\
\bottomrule[1.2pt]
\end{tabular}%
}
\vspace{-5pt}
\caption{Evaluation on image benchmarks for LLaVA-1.5-13B under different token reduction ratio.}
\vspace{-5pt}
\label{tab:app_llava_13b_benchmarks}
\end{table*}

%% file: Appendix/further_ablation.tex
\begin{table}[!ht]
\centering
\scriptsize
\setlength{\tabcolsep}{5pt}
\renewcommand{\arraystretch}{1.0}
\footnotesize

\scalebox{0.75}{
\begin{tabular}{l ccccc}
\toprule[1.2pt]

\textbf{Method} & \textbf{GQA} & \textbf{MME} & \textbf{POPE} & \textbf{VQA\textsuperscript{Text}} & \textbf{Avg.} \\
\midrule

\rowcolor{gray!10} \textit{Upper Bound (576 Tokens)} & 61.9 & 1511 & 85.9 & 58.2 & 100.0\% \\

\midrule

\rowcolor{gray!10} \textbf{LLaVA-1.5-7B} & \multicolumn{4}{c}{\textit{Retain 32 Tokens} \textbf{\textcolor{mydarkgreen}{($\downarrow$ 94.4\%)}}} & \multicolumn{1}{c}{} \\
DivPrune\confyear{CVPR25} & 55.0 & 1306 & 81.7 & 53.1 & 90.4\% \\

\midrule
\textit{Centering} & \multicolumn{5}{c}{} \\

No Centering & 55.4 & 1329 & 81.4 & 55.0 & 91.7\% \\
Mean-Direction Removal & 55.6 & 1322 & 81.3 & 54.8 & 91.5\% \\
\rowcolor{mysoftgreen} Mean-Centering (Ours) & 56.0 & 1335 & 82.8 & 55.3 & \textbf{92.6\%} \\
\midrule
\textit{Distinctiveness Variants} & \multicolumn{5}{c}{} \\


\texttt{[CLS]} Attention & 55.0 & 1306 & 79.3 & 54.5 & 90.3\% \\

Raw-$\ell_2$ & 55.9 & 1338 & 83.3 & 53.8 & 92.0\% \\

\rowcolor{mysoftgreen} Raw-Cosine (Ours) & 56.0 & 1335 & 82.8 & 55.3 & \textbf{92.6\%} \\




\midrule
\textit{Distinctiveness Normalization} & \multicolumn{5}{c}{} \\
Z-Norm. + Sigmoid & 56.0 & 1325 & 83.6 & 54.6 & 92.3\% \\
\rowcolor{mysoftgreen} Min--Max Norm. (Ours) & 56.0 & 1335 & 82.8 & 55.3 & \textbf{92.6\%} \\

\bottomrule[1.2pt]
\end{tabular}
}
\vspace{-5pt}
\caption{Further ablation studies on LLaVA-1.5-7B.}
\vspace{-10pt}
\label{tab:app_further_ablation}
\end{table}

%% file: Appendix/Efficiency_llava-1.5-7b.tex
\begin{table}[!h]
    \centering
    

\setlength{\tabcolsep}{5pt} 
\renewcommand{\arraystretch}{0.95} 

\footnotesize
\scalebox{0.75}{
\begin{tabular}{l  ccc}
\toprule[1.2pt]
\textbf{Method} & \textbf{Prefill (ms)} & \textbf{Latency (ms)} & \textbf{FLOPs (G)} \\
\midrule
\rowcolor{gray!10} \textit{Upper Bound (576 Tokens)}  & 154.77 & 200.27 & 4431.21 \\
\midrule
\rowcolor{gray!10} \textbf{LLaVA-1.5-7B} & \multicolumn{3}{c}{\textit{Retain 128 Tokens} \textbf{\textcolor{mydarkgreen}{($\downarrow$ 77.8\%)}}} \\ 

DivPrune\confyear{CVPR25} & 93.50 & 136.65 & 1471.24 \\
\rowcolor{mysoftgreen} \quad + \textbf{Ours}  & 94.51 & 137.10 & 1471.24 \\
CDPruner\confyear{NeurIPS25}  & 118.45 & 161.27 & 1479.61\\
\rowcolor{mysoftgreen} \quad + \textbf{Ours} & 119.60 & 162.20 & 1479.61 \\
ZOO-Prune\confyear{CVPR26}  & 129.99 & 172.21 & 3018.78 \\
\rowcolor{mysoftgreen} \quad + \textbf{Ours}  & 130.20 & 172.94 & 3018.78 \\

\midrule
\rowcolor{gray!10} \textbf{LLaVA-1.5-7B} & \multicolumn{3}{c}{\textit{Retain 64 Tokens} \textbf{\textcolor{mydarkgreen}{($\downarrow$ 88.9\%)}}} \\ 

DivPrune\confyear{CVPR25}  & 77.68 & 119.53 & 1048.38 \\
\rowcolor{mysoftgreen} \quad + \textbf{Ours}  & 77.82 & 119.85 & 1048.38 \\
CDPruner\confyear{NeurIPS25}  & 90.92 & 133.54 & 1056.75 \\
\rowcolor{mysoftgreen} \quad + \textbf{Ours}  & 91.85 & 134.63 & 1056.75 \\
ZOO-Prune\confyear{CVPR26}  & 112.08 &154.51 & 2595.93 \\
\rowcolor{mysoftgreen} \quad + \textbf{Ours}  & 112.32 & 154.77 & 2595.93 \\

\midrule
\rowcolor{gray!10} \textbf{LLaVA-1.5-7B} & \multicolumn{3}{c}{\textit{Retain 32 Tokens} \textbf{\textcolor{mydarkgreen}{($\downarrow$ 94.4\%)}}} \\ 

DivPrune\confyear{CVPR25}  & 69.38 & 111.22 & 836.96 \\
\rowcolor{mysoftgreen} \quad + \textbf{Ours}  & 69.48 & 111.07 & 836.96 \\
CDPruner\confyear{NeurIPS25}  & 75.57 & 118.03 & 845.32 \\
\rowcolor{mysoftgreen} \quad + \textbf{Ours}  & 75.67 & 118.53 & 845.32 \\
ZOO-Prune\confyear{CVPR26}  & 105.82 & 148.32 & 2348.51 \\
\rowcolor{mysoftgreen} \quad + \textbf{Ours}  & 105.94 & 148.59 & 2348.51 \\

\bottomrule[1.2pt]
\end{tabular}}
    \vspace{-5pt}
    \caption{Efficiency comparison on LLaVA-1.5-7B.}
    \label{tab:app_efficiency_complete}
    \vspace{-10pt}
\end{table}